\documentclass{article} 
\usepackage[preprint]{colm2026}

\usepackage{arydshln}
\usepackage{caption}
\usepackage{microtype}
\usepackage{hyperref}
\usepackage{url}

\usepackage{amsfonts}       %
\usepackage{nicefrac}       %
\usepackage[table]{xcolor}         %
\usepackage{mdframed}
\usepackage{xspace}
\usepackage{enumitem}
\usepackage{subcaption} 
\usepackage{booktabs, multirow, makecell}

\usepackage{tcolorbox}
\usepackage{soul}
\tcbuselibrary{skins, listings, breakable}
\usepackage[T1]{fontenc}
\usepackage{longtable}
\usepackage{threeparttable}
\usepackage{tabularx}
\usepackage{pifont}
\usepackage{amsmath} 
\usepackage{cleveref}
\usepackage{wrapfig}
\newcommand{\nmark}{--}
\usepackage{array}
\usepackage{pifont}

\newcommand{\cmark}{\ding{51}}

\newcolumntype{Y}{>{\centering\arraybackslash}X}

\newcommand{\orbit}[0]{ORBIT\xspace}

\definecolor{odysseyblue}{HTML}{5A63B7}
\definecolor{odysseyrow}{HTML}{EEF0FA} 

\definecolor{hlblue}{RGB}{198,217,241}   
\definecolor{hlpurple}{RGB}{221,217,238} 
\definecolor{hlgreen}{RGB}{226,239,218}  
\definecolor{hlorange}{RGB}{255,230,204} 
\definecolor{hlyellow}{RGB}{255,242,204} 
\definecolor{paperblue}{RGB}{0, 114, 178}
\definecolor{orb_thinkcolor}{RGB}{0, 114, 178}   
\definecolor{orb_searchcolor}{RGB}{0, 153, 136}  
\definecolor{orb_infocolor}{RGB}{238, 119, 51}   
\definecolor{orb_anscolor}{RGB}{204, 51, 17}      

\newcommand{\orbtagthink}[1]{%
  \textcolor{orb_thinkcolor}{\texttt{<think>}} #1 \textcolor{orb_thinkcolor}{\texttt{</think>}}}
\newcommand{\orbtagsearch}[1]{%
  \textcolor{orb_searchcolor}{\texttt{<search>}} \emph{#1} \textcolor{orb_searchcolor}{\texttt{</search>}}}
\newcommand{\orbtaginfo}[1]{%
  \textcolor{orb_infocolor}{\texttt{<information>}} #1 \textcolor{orb_infocolor}{\texttt{</information>}}}
\newcommand{\orbtagans}[1]{%
  \textcolor{orb_anscolor}{\texttt{<answer>}} \textbf{#1} \textcolor{orb_anscolor}{\texttt{</answer>}}}

\newcommand{\Hblue}[1]{\sethlcolor{hlblue}\hl{#1}} 
\newcommand{\Horange}[1]{\sethlcolor{hlorange}\hl{#1}}
\newcommand{\Hgreen}[1]{\sethlcolor{hlgreen}\hl{#1}} 
\newcommand{\Hpurple}[1]{\sethlcolor{hlpurple}\hl{#1}} 
\newcommand{\Hyellow}[1]{\sethlcolor{hlyellow}\hl{#1}}

\crefformat{section}{\S#2#1#3} 
\crefformat{subsection}{\S#2#1#3}
\crefformat{subsubsection}{\S#2#1#3}

\DeclareCaptionLabelFormat{subfig}{Figure~\thefigure\alph{subfigure}}
\definecolor{darkblue}{rgb}{0, 0, 0.5}
\hypersetup{colorlinks=true, citecolor=darkblue, linkcolor=darkblue, urlcolor=darkblue}

\usepackage{lineno}

\title{%
  \raisebox{-0.25\height}{\includegraphics[height=0.8cm]{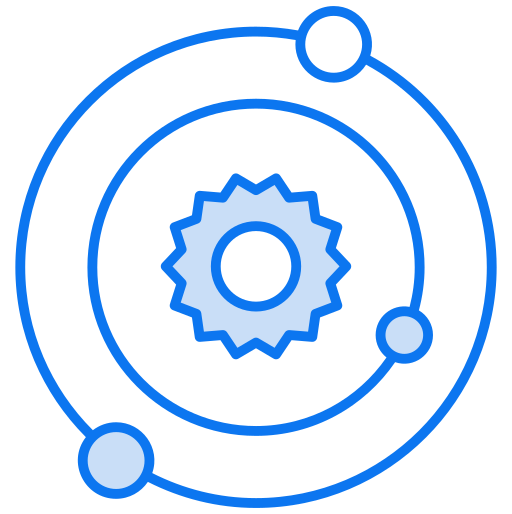}}%
  ~\textcolor{paperblue}{ORBIT}: Scalable and Verifiable Data Generation for Search Agents on a Tight Budget
}

\author{Nandan Thakur, Zijian Chen, Xueguang Ma \& Jimmy Lin \\
David R. Cheriton School of Computer Science \\
University of Waterloo \\
\texttt{\{n3thakur,s42chen,x93ma,jimmylin\}@uwaterloo.ca} \\
\And
}

\begin{document}

\ifcolmsubmission
\linenumbers
\fi

\maketitle
\ifcolmpreprint
  \lhead{Preprint. Under review.}
\fi

\begin{abstract}
Search agents, which integrate language models (LMs) with web search, are becoming crucial for answering complex user queries. 
Constructing training datasets for deep research tasks, involving multi-step retrieval and reasoning, \emph{remains challenging} due to expensive human annotation, or cumbersome prerequisites.
In this work, we introduce \textbf{\orbit}, a training dataset with 20K reasoning-intensive queries with short verifiable answers, generated using a \emph{frugal framework} without relying on paid API services. 
The modular framework relies on four stages: seed creation, question–answer pair generation, and two stages of verification: self and external. 
\orbit spans 15 domains and each training pair requires 4--5 reasoning steps, with external search verification required from the complete web. 
We train Qwen3-4B as the base model on \orbit using GRPO and evaluate it on Wikipedia question answering tasks. Extensive experiment results demonstrate that \orbit-4B achieves strong performance among sub-4B LLMs as search agents, proving the utility of synthetic datasets. 
Our framework, code and datasets are open-sourced and available publicly.
\end{abstract}

\vspace{-3mm}
\begin{center}
\newcommand{\logoh}{1.35em}

\href{https://github.com/castorini/orbit}{
\raisebox{-0.2\height}{\includegraphics[height=\logoh]{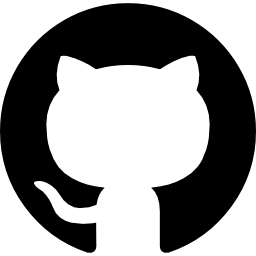}}
\hspace{0.35em}\textcolor{paperblue}{\textbf{\texttt{castorini/orbit}}}
}
\quad
\href{https://huggingface.co/orbit-ai}{
\raisebox{-0.2\height}{\includegraphics[height=\logoh]{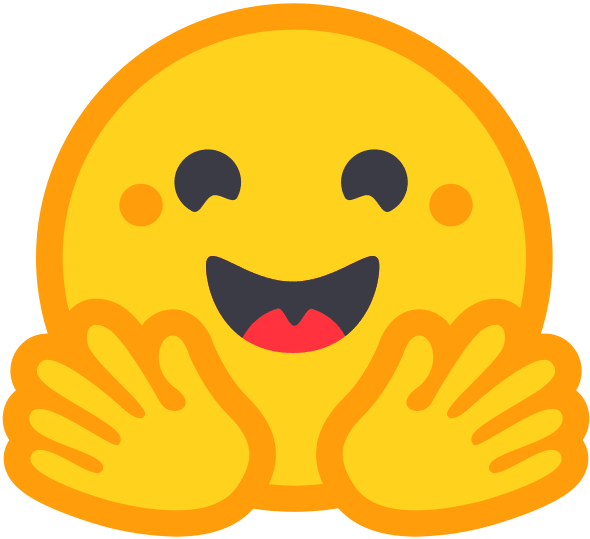}}
\hspace{0.35em}\textcolor{paperblue}{\textbf{\texttt{orbit-ai}}}
}
\end{center}
\vspace{-2mm}


\section{Introduction}

Large language models (LLMs) have been used for their reasoning capabilities beyond simple factoid queries in a new frontier we refer to as \emph{deep search}. This setting requires interleaved reasoning, decomposing complex tasks, and reasoning across multiple sources of information, and it is now exposed as an endpoint in multiple commercial products~\citep{openai2025deepresearch, gemini2025deepresearch}. Unlike traditional question-answering systems, answering \emph{deep search} questions goes beyond a single-turn web search: it requires breaking down complex queries, conducting multiple retrieval steps, and iteratively planning searches and aggregating results from a search tool. 
Recent work training search agents, such as Search-R1 and InfoSeek~\citep{search_r1, infoseek} has shown that applying GRPO with verifiable rewards improves models’ iterative reasoning and search tool use.

Training data availability for \emph{deep search} tasks is very limited. Search-R1 trains on NQ~\citep{kwiatkowski-etal-2019-natural} and HotpotQA~\citep{yang-etal-2018-hotpotqa} datasets; however, these contain simple retrieval queries that do not necessarily require multiple hops of searches for answering them. 
Collecting human annotations with verifiable answers for complex queries is also challenging: given a complex query, the uncertainty of web search makes the annotation burden impractically high. 
On the other hand, training datasets with complex search queries are either limited in training pairs, or require pre-requisites before generation, such as knowledge-graph construction, limiting their real-world applicability and usage~\citep{webwalker, simpledeepsearcher, infodeepseek, webshaper, webexplorer, monaco, infoseek}.

In this work, we introduce a frugal framework for constructing complex training data for search agents in a budget constrained manner, without relying on any paid API services. 
The framework is modular and can be adopted easily for synthetic data generation.
Existing synthetic deep search datasets commonly rely on graph structures and entity linkages, or are not fully open sourced, as shown in \autoref{tab:dataset-comparison}. Using our framework, we construct \textbf{\orbit} (\emph{Open-Web Reasoning for Information Retrieval Tasks}), a verified synthetic training dataset with narrative-style, i.e., long and reasoning-intensive questions with short verifiable answers, requiring multi-hops across the web.

Specifically, \orbit is constructed by a four-stage framework that automatically constructs verifiable, multi-hop training pairs at scale: (1) \emph{seed creation}: we expand 15 domains into large sets of Wikipedia categories and iterate linked pages, using page titles as seeds to ensure both head and long-tail coverage; (2) \emph{question--answer pair generation}: conditioned on a seed, a search-enabled \texttt{DeepSeek-V3.1} composes an inverted question paired with a short verifiable answer; (3) \emph{self-verification}: a search-enabled \texttt{DeepSeek-V3.1} assesses whether the proposed answer satisfies every atomic fact mentioned in the inverted question and provides supporting URLs as citations, discarding pairs with unverified answers or weakly supported sub-claims; and (4) \emph{external-verification}: two cascading LLM judges (\texttt{Qwen3-4B-Instruct-2507} and \texttt{gpt-oss-120b}) with scraped URL context, generate an answer and self-verify the generated answer as the judge with the ground-truth answer.

Using \orbit, we fine-tune smaller and efficient LLMs (<4B parameters) as search agents using Qwen3-4B as the base model~\citep{qwen3}, with GRPO~\citep{shao:2024}. By mixing \orbit with training datasets, \orbit-4B demonstrates strong effectiveness over sub-4B search agent baselines on complex Wikipedia-based QA benchmarks, highlighting the potential of our framework for data generation to produce high-quality supervision for scaling small-sized search agents. We summarize our major contributions as follows:

\begin{enumerate}[leftmargin=*]
    \item We propose a frugal construction framework without requiring any pre-requisites or relying on expensive API's, to generate reasoning-intensive queries with short and verifiable answers on a tight budget.
    \item We introduce \orbit, one of the largest available datasets for training search agents with 20K+ verified question--answer pairs with a hard complexity, requiring verification from both Wikipedia and Web domains for each question.
    \item We empirically demonstrate the effectiveness of \orbit through extensive experiments, showing that the proposed dataset yields significant performance gains on small-sized search agents. \orbit-4B surpasses existing search agents with less than 4B model parameters by up to 9.0 EM accuracy on complex QA benchmarks on Wikipedia. 
    \item We open-source the framework, \orbit dataset and the code for training search agents. Code is available on GitHub at: \url{https://github.com/castorini/orbit}, \orbit dataset and search agents are available on Hugging Face at: \url{https://huggingface.co/orbit-ai}.
\end{enumerate}

\newcommand{\dscite}[2]{#1\,\citep{#2}}

\begin{table}[t]
\centering

{\setlength{\tabcolsep}{2pt}
\renewcommand{\arraystretch}{0.8}
\fontsize{8.5}{12}\selectfont

\begin{tabularx}{\linewidth}{@{}
  >{\raggedright\arraybackslash}p{0.30\linewidth}
  >{\centering\arraybackslash}p{0.12\linewidth}
  >{\centering\arraybackslash}p{0.11\linewidth}
  >{\centering\arraybackslash}p{0.13\linewidth}
  >{\raggedright\arraybackslash}X
@{}}
\toprule
\textbf{Dataset Name} & \textbf{Source} & \textbf{Train Pairs} & \textbf{Complexity} & \textbf{Pre-requisite / Constraint} \\
\midrule
\dscite{NQ}{kwiatkowski-etal-2019-natural} & Wiki & 300K+ & Easy & None \\
\dscite{HotpotQA}{yang-etal-2018-hotpotqa} & Wiki & 100K+ & Easy & None \\
\midrule
\dscite{WebWalkerQA}{webwalker} & Web & 14.3K & Easy & Web Traversal \\
\dscite{SimpleDeepS.}{simpledeepsearcher} & Wiki~\&~Web & 871 & Easy & Web Search \& HTML Extract \\
\dscite{InfoDeepSeek}{infodeepseek} & Web & 245 & Easy & Human Annotators \\
\dscite{WebShaper}{webshaper} & Wiki & 500 & Hard & Wikipedia Hyperlinks \\
\dscite{WebExplorer}{webexplorer} & Web & 100 & Hard & Web Search \& Obfuscation \\
\dscite{MoNaCo}{monaco} & Wiki & 1,315 & Hard & Human Annotators \\
\dscite{DeepDive}{deepdive} & KG & 3,250 & Hard & Knowledge Graph \\
\dscite{InfoSeek}{infoseek} & Wiki~\&~Web$^\dagger$ & 17.8K & Hard & Entity Linking \& Obfuscation \\
\midrule

\rowcolor{odysseyrow} \textbf{\textcolor{paperblue}{\orbit\ (ours)}}
& \textcolor{paperblue}{\textbf{Wiki \& Web}} & \textcolor{paperblue}{\textbf{20K}} & \textcolor{paperblue}{\textbf{Hard}} & \textcolor{paperblue}{\textbf{None}} \\
\bottomrule
\end{tabularx}
}
\caption{Comparison of \orbit against other training datasets for search agents. 
Prior datasets require prior constraints such as a knowledge graph (KG), lack structural depth or remain limited in scale. \orbit is a large-scale multi-hop dataset, requiring no prior constraints and containing complex queries, that can be scaled easily with a low budget.}
\vspace*{-\baselineskip}
\label{tab:dataset-comparison}
\end{table}


\vspace{-3mm}
\section{Related Work}

\subsection{Reinforcement Learning and Search-Augmented LLMs}
Reinforcement learning (RL) improves agent performance by learning from previous experience and maximizing cumulative rewards~\citep{sutton:1998}. Recently, RL with group-based methods such as GRPO~\citep{shao:2024} have enabled LLMs to solve complex tasks such as Olympiad-level math problems as well as broader reasoning tasks~\citep{shao:2024, yu2025dapo, wen2026reinforcement}. On the search side, retrieval-augmentation improves LLM by integrating external knowledge~\citep{khandelwal:2020,lewis:2020,gao-etal-2023-enabling}. Furthermore, iteratively retrieving relevant documents improves LLM performance on complex questions~\citep{jiang-etal-2023-active, trivedi-etal-2023-interleaving, mallen-etal-2023-trust, yoran2024making}. In search agents, search engines are popularly incorporated as optional tools, leading to enhanced multi-hop reasoning performance, such as Search-R1, Search-o1, among many others~\citep{search_r1, r1_searcher, search_o1, deepresearcher}. We provide a comprehensive summary of search agents in \autoref{sec:extended-related-work}.

\begin{table}[t]
    \centering
    {\setlength{\tabcolsep}{6.5pt}
    \renewcommand{\arraystretch}{0.8}
    \fontsize{8.2}{12}\selectfont
    \begin{tabular}{@{} p{\linewidth} @{}}
    \toprule
    \textbf{Question:} What was the exact runtime (minutes) of the \Hblue{2017 animated feature set inside a smartphone's messaging application}, \Horange{directed by a filmmaker previously known for sequels to popular children's franchises}, \Hpurple{featuring a protagonist whose facial expression malfunctions}, \Hgreen{with voice casting that includes a lead actor from a critically acclaimed sitcom}, and \Hyellow{produced by a studio that later won an Oscar for Spider-Man animation}? \\
    \textbf{Answer: 86 minutes} \\ \midrule
    \rowcolor{odysseyrow} \multicolumn{1}{l}{{\textbf{Verification~/~Summary of Supporting Facts:}}} \\
    \textcolor[HTML]{336600}{\cmark} \Hblue{\textbf{Animated Set}:} \emph{\textcolor{blue}{The Emoji Movie} released in 2017, set inside a smartphone messaging app world called Textopolis.} \\
    \textcolor[HTML]{336600}{\cmark} \Horange{\textbf{Filmmaker:}} \emph{\textcolor{blue}{Tony Leondis}, sequels to franchises: Lilo \& Stitch 2, Kung Fu Panda: Secrets of the Masters}. \\
    \textcolor[HTML]{336600}{\cmark} \Hpurple{\textbf{Protagonist}}: \emph{\textcolor{blue}{Gene}, the main character in The Emoji Movie, struggles with malfunctioning expressions.} \\
    \textcolor[HTML]{336600}{\cmark} \Hgreen{\textbf{Voice Cast~:}} \emph{\textcolor{blue}{T.J. Miller}, who was a lead actor from the HBO sitcom titled Silicon Valley.} \\
    \textcolor[HTML]{336600}{\cmark} \Hyellow{\textbf{Voice Casting Studio~:}} \emph{\textcolor{blue}{Sony Pictures Animation}, that won an Oscar for Spider-Man: Into the Spider-Verse.} \\
    \bottomrule
    \end{tabular}
    }
    \vspace{-2mm}
    \caption{A training data example showing the question, answer and verification (for the reader) from the \emph{TV Shows \& Movies} domain in the \orbit dataset. Colored spans mark the distinct factual clues embedded in the question; the verification rows confirm each clue with key supporting evidence (\textcolor{blue}{blue}). More dataset examples are provided in \autoref{app:domain-examples}.}
    \label{tab:odyssey-example}
    \end{table}

\subsection{Comparison with Existing Search Agent Training Datasets}
Designing complex user queries with verifiable answers for deep research is nontrivial, because these tasks inherently require extensive search and exploration to find the correct answer. 
This limits the scale of human-annotated evaluation data, since verifying correctness and grounding answers to evidence documents is time-consuming.
Synthetic data generation with large language models offers a way to scale training data, as shown in \autoref{tab:dataset-comparison}. 
Available datasets such as NQ~\citep{kwiatkowski-etal-2019-natural} and HotpotQA~\citep{yang-etal-2018-hotpotqa}, commonly used for finetuning search agents, such as Search-R1, contain easy queries, solvable within 1--2 hops of search. 

This motivated works on constructing deep search datasets with complex queries. Existing works such as DeepDive~\citep{deepdive}, WebShaper~\citep{webshaper} and WebSailor~\citep{websailor} typically rely on hyperlink structure or knowledge graphs to synthesize queries from graph context. Similarly, works such as MoNaCo~\citep{monaco} and InForage~\citep{inforage} rely on human annotators, whereas InfoSeek~\citep{infoseek}, a concurrent effort constructs synthetic datasets via entity linking and obfuscation. 
Nevertheless, existing methods primarily include cumbersome pre-requisites that inherently limit the reproduction or real-world adoption. In contrast, we target a strict \emph{no prerequisites} setting, relying exclusively on search-enabled LLMs to provide us with synthetic QA training pairs with reasonable complexity (an example in \autoref{tab:odyssey-example}) without requiring any paid API services. In terms of scale, \orbit contains 20K training pairs, the largest amongst concurrent training datasets containing queries with hard complexity, covering both Wikipedia and Web (Britannica, NIH, StackExchange, etc.) as sources on 15 diverse domains.


\section{\orbit: A Scalable and Verifiable Training Dataset for Search Agents}

\orbit consists of 20K \emph{reasoning-intensive} or complex queries, often requiring \emph{multi-hops} of searching information to be able to answer them confidently. We carefully constructed the dataset through a fully-automatic four-stage framework, using seeds constructed across 15 diverse domains. We first describe the data collection pipeline in Section \ref{sec:dataset-creation}, then analyze the dataset statistics in Section \ref{sec:dataset-stats}. 

\subsection{\orbit Dataset Creation}\label{sec:dataset-creation}

\orbit dataset creation requires zero pre-requisites allowing researchers to easily adopt the framework. Each training pair in \orbit includes an \emph{inverted} question and \emph{short verifiable answer} (example in \autoref{tab:odyssey-example}), where it is easier to verify the question given the answer~\citep{browsecomp:2025}. The dataset construction framework comprises four stages, described below:

\paragraph{\textcolor[HTML]{5A4E8C}{Stage 1: Seed Creation.}} We manually decide on a set of 15 broad domains to be used, 12 of these broad domains are inspired from BrowseComp~\citep{browsecomp:2025}: \emph{TV Shows \& Movies}, \emph{Science \& Technology}, \emph{Art}, \emph{History}, \emph{Sports}, \emph{Music}, \emph{Video Games}, \emph{Geography}, \emph{Politics}, \emph{Medicine}, \emph{Finance} and \emph{Law}. We add three more: \emph{Puzzles}, \emph{Mathematics} and \emph{Code}. Next, using OpenAI's \textcolor[HTML]{5A4E8C}{\texttt{Deep Research}} (\texttt{\href{https://chat.openai.com/}{chat.openai.com}}), we generate at least 100 different Wikipedia categories for each domain that are important, with a focus on diversity covering both frequent (head) and tail-end categories. We defer to the complete list of categories for each domain in \autoref{tab:openai-categories}. 
Furthermore, we use MediaWiki's REST API\footnote{\url{https://www.mediawiki.org/wiki/API:REST_API}} to retrieve Wikipedia pages that are linked to a particular Wikipedia category. We use the linked Wikipedia page titles as \textbf{seeds} during question--answer pair generation in the next stage.

\begin{figure*}[t]
  \centering
  \includegraphics[width=\textwidth, trim=10 10 10 5, clip]{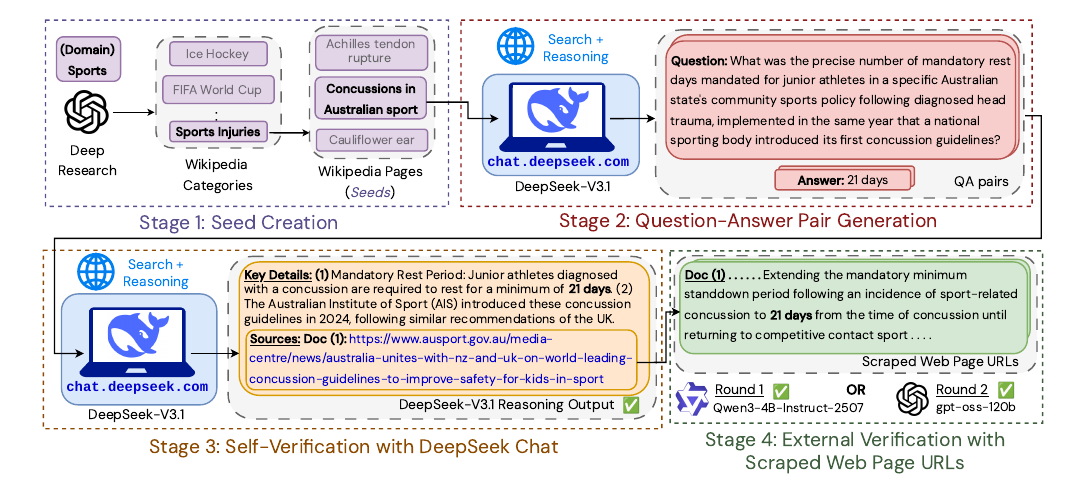}
  \vspace{-1.5em}
    \caption{An end-to-end data instance in the \textbf{\orbit} dataset generated using our framework. The procedure involves the following stages: (1) \emph{seed creation}, (2) \emph{question--answer pair generation}, (3) \emph{self-verification with DeepSeek Chat}, and (4) \emph{external verification with scraped web page URLs}. Each stage is described in detail in Section \ref{sec:dataset-creation}.
    }
  \label{fig:odyssey-flowchart}
\end{figure*}

\paragraph{\textcolor[HTML]{8B1A1A}{Stage 2: Question--Answer Pair Generation.}} We utilize the Wikipedia page title as the seed as an inspiration to generate an \emph{inverted} question, that is \emph{reasoning-intensive}, requiring multi-hop verifications to reach the generated answer. 
Specifically, we utilize \textcolor[HTML]{8B1A1A}{\texttt{DeepSeek-V3.1}} by sending each prompt on DeepSeek's Chat GUI (\texttt{\href{https://chat.deepseek.com/}{chat.deepseek.com}}) automatically using Python Selenium. 
We require a manual login to the website once, and activate the \texttt{DeepThink} \& \texttt{Search} buttons. 
\textcolor[HTML]{8B1A1A}{\texttt{DeepSeek-V3.1}} conducts a web-search, gathers information, reasons and thinks carefully by checking each information hop, and generates the question--answer pair in the required output format. 
The prompt includes a single-shot examplar (shuffled for answer diversity\footnote{We empirically observed that shuffling the exemplar helped generate question--answer pairs with diverse types of answers, e.g., not always fixated on the year or a numerical answer.}) and is provided in \autoref{fig:prompt-multihop}. We generated a total of 44.1K training pairs from Stage 2 with \textcolor[HTML]{8B1A1A}{\texttt{DeepSeek-V3.1}}, requiring 2--3 months for the completion of this stage.

\paragraph{\textcolor[HTML]{8C4A00}{Stage 3: Self-Verification with DeepSeek Chat.}} After generating synthetic training pairs from the previous stage, we conduct \emph{self-verification} for the question--answer pair. 
For this stage, we require a high-performing LLM with access to web search; however, closed-source search APIs are expensive. 
Therefore, to limit costs, we independently prompt the same model: \textcolor[HTML]{8C4A00}{\texttt{DeepSeek-V3.1}} from the DeepSeek Chat GUI (\texttt{\href{https://chat.deepseek.com/}{chat.deepseek.com}}) with enabled \texttt{DeepThink} and \texttt{Search} buttons, by providing the generated question and the answer as input, and ask \textcolor[HTML]{8C4A00}{\texttt{DeepSeek-V3.1}} to reason and verify whether the answer satisfies all criteria and sub-parts mentioned in the question. The self-verification prompt is shown in \autoref{fig:prompt-self-verification}. DeepSeek Chat first conducts a web-search by searching a maximum of 50 documents (before starting to think) to be able to verify and cite the required sources as verification statements. Once completed, we provide the \textcolor[HTML]{8C4A00} reasoning output to \textcolor[HTML]{8C4A00}{\texttt{Qwen3-4B-Instruct-2507}}\footnote{\url{https://huggingface.co/Qwen/Qwen3-4B-Instruct-2507}} to filter out training pairs with answers without sufficient verification, i.e., fully incorrect, or those only partially answered the question. After Stage 3, we were able to self-verify 61.5\% of all training pairs from Stage 2, remaining with 27.1K training pairs.

\paragraph{\textcolor[HTML]{3F5F3F}{Stage 4: External Verification with Scraped Web Page URLs.}} To avoid self bias using \texttt{DeepSeek-V3.1} for both pair generation and verification~\citep{panickssery2024llm}, we utilize two separate LLMs in a cascading setup to conduct an external verification.
We scrape the web context available from the web page URLs provided during the self-verification stage using Python Selenium, and parse the document with Trafilatura~\citep{barbaresi-2021-trafilatura}.
In the first round, we prompt the \textcolor[HTML]{3F5F3F}{\texttt{Qwen3-4B-Instruct-2507}} judge by providing the question, answer and scraped web page document information, and ask the LLM to independently reason and provide the answer. Next, we use the same LLM to evaluate whether the LLM judge answer and the gold answer match. The external verification prompt is provided in \autoref{fig:prompt-external-generation}. For all unmatched pairs in the first round, we repeat the procedure with OpenAI's \textcolor[HTML]{3F5F3F}{\texttt{gpt-oss-120b}}\footnote{\url{https://huggingface.co/openai/gpt-oss-120b}} as the judge in the next round. 
After external verification from both rounds, we verify 19,790 training pairs. 
Lastly, the authors manually verify and add 210 training pairs by checking the non-verified 400--500 training pairs by interacting with OpenAI's ChatGPT (\texttt{\href{https://chat.openai.com/}{chat.openai.com}}) (GPT-5.2 with \texttt{Search} button enabled).

\begin{figure}[t]
  \centering
    \begin{subfigure}[t]{0.47\textwidth}
    \includegraphics[width=\textwidth, trim=40 40 10 40, clip]{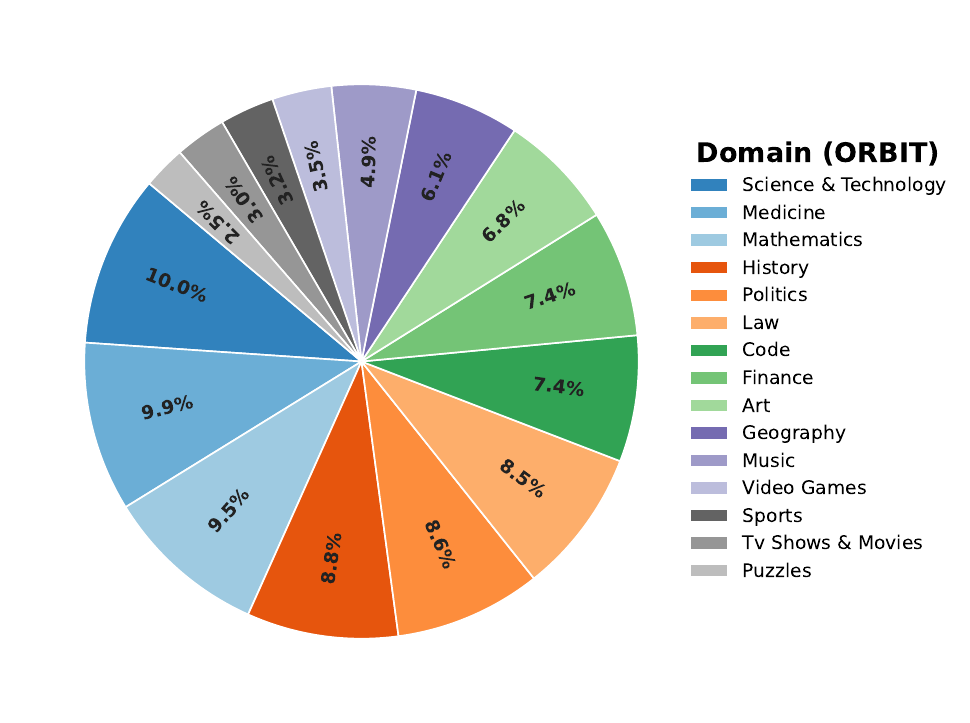}
    \caption{\textbf{\orbit domain composition.} We have question--answer pairs distributed evenly across 15 domains. The highest number of pairs come from \emph{Science \& Technology} (10\%) and lowest from \emph{Puzzles} (2.5\%).
    }
    \label{fig:domain-composition}
  \end{subfigure}
  \hfill
  \begin{subfigure}[t]{0.47\textwidth}
    \includegraphics[width=\textwidth]{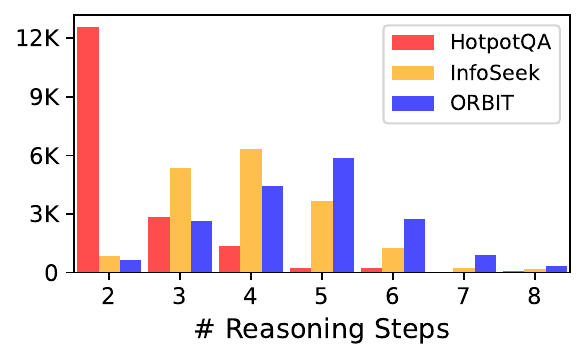}
    \caption{\textbf{Question complexity.} \orbit contains complex or deep search queries requiring on average 4--6 reasoning steps, that is higher than existing datasets: InfoSeek (3--5 steps) and HotpotQA (2 steps).
    }
    \label{fig:hops-distribution}
  \end{subfigure}
  \vspace{-1em}
\end{figure}

\begin{figure}[t]
  \centering
  \begin{subfigure}[t]{0.48\textwidth}
    \includegraphics[width=\textwidth, trim=5 0 10 0, clip]{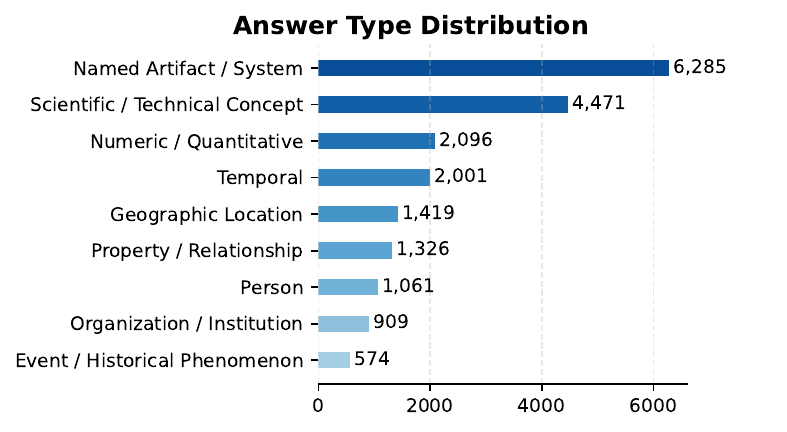}
    \caption{\textbf{Answer type distribution.} Answers are a mixture of named artifacts, scientific or technical concepts, or even historical phenomenon. Answer type definitions are provided in \autoref{tab:answer-types}.
    }
    \label{fig:answer-type-distribution}
  \end{subfigure}
  \hfill
  \begin{subfigure}[t]{0.46\textwidth}
    \includegraphics[width=\textwidth, trim=0 0 0 0, clip]{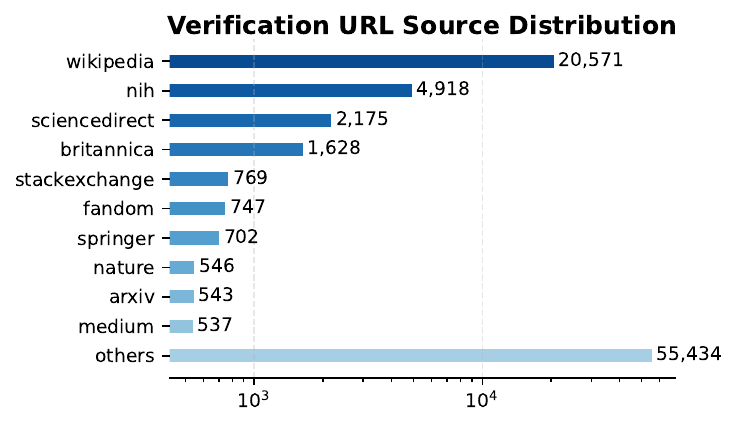}
    \caption{\textbf{URL source distribution.} Verification URL's from the head distribution include Wikipedia, NIH, Britannica among others. There is a long-tail of source distribution.
    }
    \label{fig:url-source-distribution}
  \end{subfigure}
  \vspace{-1em}
\end{figure}

\subsection{\orbit Dataset Statistics}\label{sec:dataset-stats}

The \orbit dataset statistics are provided in \autoref{tab:odyssey-stats}. Generated questions are long, containing 64 tokens on average and answers are short (and verifiable) with 3.5 tokens (measured using OpenAI's tiktoken package). The average number of estimated reasoning steps (by counting the number of verification statements) is 4.42, thereby requiring 4--5 information hops to verify the answer. Similarly, there are about 4.36 verification URLs on average, with almost 3.35$\times$ of non-Wikipedia URLs over Wikipedia URLs.

\paragraph{Domain Composition.} We show the domain composition present in \orbit in \autoref{fig:domain-composition}.  
Questions from each domain in \orbit are generated uniformly and filtered by removing unverified question--answer pairs. 
\emph{Science \& Technology} has the highest composition of question--answer pairs with 10\% and \emph{Puzzles} has the lowest composition with 2.5\%. To summarize, \orbit contains question--answer pairs spread evenly across all domains. 

\begin{wraptable}{r}{0.35\columnwidth}
\vspace{-15pt}  
\centering
{\setlength{\tabcolsep}{8pt}
\renewcommand{\arraystretch}{0.95}
\fontsize{8}{12}\selectfont

\begin{tabular}{@{} l r @{}}
\toprule
\textbf{Dataset Statistic} & \textbf{Value} \\
\midrule
\textbf{\# Question--Answer pairs} & 20,147 \\
\quad - Avg. \# question tokens & 63.87 \\
\quad - Avg. \# answer tokens & 3.46 \\
\quad - Avg. \# reasoning steps & 4.42 \\ \midrule
\textbf{Avg. Verification URLs} & 4.36 \\
\quad - Non-Wikipedia URLs & 3.35 \\
\quad - Wikipedia URLs & 1.02 \\
\bottomrule
\end{tabular}
}
\caption{\textbf{\orbit dataset statistics}. 
URL statistics reflect the distinct sources required for answer verification.}
\label{tab:odyssey-stats}
\vspace{-8pt}
\end{wraptable}

\paragraph{Question Complexity.} We evaluate the amount of information-seeking or reasoning hops necessary to faithfully answer each part of the question. We work on three datasets: (1) HotpotQA~\citep{yang-etal-2018-hotpotqa}, (2) InfoSeek~\citep{infoseek} and \orbit. To avoid the uneven distribution of training pairs, we randomly sample 18K training pairs from each dataset. We decompose the original question into necessary sub-questions with Qwen3-8B~\citep{qwen3} (with thinking mode enabled), and count each sub-question as a separate information hop \emph{necessary} to verify a part of the question. Our results in \autoref{fig:hops-distribution}, show that \orbit contains the most complex queries, requiring on average 4--6 reasoning steps to answer them. In contrast, InfoSeek contains 3--5 reasoning steps and HotpotQA with 2 reasoning steps.

\paragraph{Answer Type and URL Source Distribution.} In \autoref{fig:answer-type-distribution}, we automatically classify the short factual answers in \orbit into each answer type category using Qwen3-8B (with thinking mode enabled). In \orbit, we observe that a majority of answers are either a named or system artifact, or a scientific or technical concept. We observe 20\% of answers based on temporal entity (e.g., calendar date, year etc.) or numeric (numerical value, measurement etc). Overall, the plot shows the diversity in \orbit answers. 
Similarly, we also plot the URL source distribution to identify major URL sources required for answer verification. From \autoref{fig:url-source-distribution}, we observe that Wikipedia is the largest source, with NIH and ScienceDirect, being the second and third-largest sources. There is a long-tail distribution of source domains (Others is the largest category) necessary for answer verification.


\begin{table*}[t]
\centering
\renewcommand{\arraystretch}{1.05}
\begin{tabularx}{\textwidth}{lYYYYYYYY}
\toprule
\multirow{2}{*}{\textbf{Model}} & \multicolumn{3}{c}{\textbf{Single-Hop QA}} & \multicolumn{4}{c}{\textbf{Multi-Hop QA}} & \multirow{2}{*}{\textbf{Avg. 7}}\\
\cmidrule(lr){2-4} \cmidrule(lr){5-8}
 & NQ & TQA & PopQA & HQA & 2Wiki & MSQ & Bamb & \\
\midrule
\rowcolor{gray!10} \multicolumn{9}{l}{\textbf{Search Agents} (Base: Qwen2.5-3B-Instruct)} \\
Search-o1-3B & 23.8 & 48.2 & 26.2 & 22.1 & 21.8 & 5.4 & 32.0 & 25.6 \\
Search-R1-3B & 40.8 & 59.1 & 42.8 & 30.8 & 31.1 & 8.4 & 13.0 & 32.3 \\
ZeroSearch-3B & 41.2 & 61.5 & 44.0 & 31.2 & 33.2 & 12.6 & 14.3 & 34.0 \\
AutoRefine-3B & 43.6 & 59.7 & 44.7 & 40.4 & 38.0 & 16.9 & 33.6 & 39.6 \\
InForage-3B & 42.1 & 59.7 & 45.2 & 40.9 & 42.8 & 17.2 & 36.0 & 40.6 \\
InfoSeeker-3B & 41.7 & 56.1 & 46.5 & \textcolor{paperblue}{\textbf{44.6}} & 50.0 & \textcolor{paperblue}{\textbf{20.5}} & 39.2 & 42.7 \\
\midrule
\rowcolor{odysseyrow} \multicolumn{9}{l}{\textcolor{paperblue}{\textbf{Search Agents} (Base: Qwen3-4B)}} \\
Search-R1-4B$^\dagger$ & 38.5 & 60.1 & 46.8 & 34.2 & 46.4 & 13.2 & 41.6 & 40.1 \\
InfoSeeker-4B$^\dagger$ & 37.0 & 64.1 & 47.0 & 39.8 & 59.0 & 15.6 & 51.2 & 44.8 \\
\cdashline{1-9}
 \rowcolor{odysseyrow} \textcolor{paperblue}{\textbf{\orbit-4B$^\dagger$}} & \textcolor{paperblue}{\textbf{43.7}} & \textcolor{paperblue}{\textbf{67.3}} & \textcolor{paperblue}{\textbf{53.8}} & 42.5 & \textcolor{paperblue}{\textbf{61.1}} & 20.1 & \textcolor{paperblue}{\textbf{55.2}} & \textcolor{paperblue}{\textbf{49.1}} \\
\bottomrule
\end{tabularx}
\caption{Performance comparison on single-hop and multi-hop QA benchmarks of search agents $<$4B parameters. 
Best scores are highlighted in \textcolor{paperblue}{\textbf{bold}}. ($\dagger$) denotes models that were trained in our work using Qwen3-4B as the base model. In addition, EM accuracy results using the 2018 Wikipedia corpus and E5-base-v2 retriever are provided in \autoref{tab:extended-comparison}.}
\vspace{-2mm}
\label{tab:results}
\end{table*}

\vspace{-2mm}
\section{Experimental Setup}
\vspace{-1mm}

\paragraph{GRPO Training Details.} 
We use the \texttt{verl-tool} repository~\citep{2025verltool} for training search agents using GRPO~\citep{shao:2024}. 
We use Qwen3-4B\footnote{\url{https://huggingface.co/Qwen/Qwen3-4B}}~\citep{qwen3} as the base model with the following hyperparameters: rollouts = 8, maximum turns = 5, maximum response length = 8192, learning rate = 1e-6, train batch size = 512, KL loss coefficient = 0.001, temperature = 1.0 and top-p = 1.0. 
The reward function is an exact match of the ground truth answer, identical to Search-R1~\citep{search_r1}.
\orbit-4B has been trained for 160 training steps, utilizing 4xH100 GPUs for 1--2 days. 
Importantly, we mix \orbit, NQ and HotpotQA in an equal mixture (1:~1:~1), as we observe that non-complex training pairs with 1--2 reasoning steps (i.e., NQ and HotpotQA) are crucial as a learning step. 
Additional details including the prompt template and full hyperparameter choices are provided in \autoref{sec:search-agent-prompt} and \autoref{sec:orbit-hyperparams}.

\paragraph{Training Search Tool.} For the search tool, we adopt the \textbf{D}ux \textbf{D}istributed \textbf{G}lobal \textbf{S}earch or \textbf{DDGS} framework, an open and free metasearch library that aggregates results from diverse web search services, including Google, Brave, Bing etc.\footnote{\url{https://github.com/deedy5/ddgs}}
The framework returns the title and a relevant snippet of the web document.
Due to budget restrictions, we are unable to use paid web search tools in our work, such as Google Search API, Exa or SerpAPI. 

\paragraph{Evaluation Setup \& Datasets.} We embed the June 2024 Wikipedia dump where articles are split into passages\footnote{\url{https://huggingface.co/datasets/Upstash/wikipedia-2024-06-bge-m3}} and retrieve top-5 passages using the BGE-M3 retrieval model~\citep{bge-m3}, replicating the evaluation setting in InfoSeek~\citep{infoseek}.
We evaluate on three single-hop Wikipedia QA benchmarks: Natural Questions (NQ)~\citep{kwiatkowski-etal-2019-natural}, TriviaQA (TQA)~\citep{joshi-etal-2017-triviaqa} and PopQA~\citep{mallen-etal-2023-trust}, and four multi-hop Wikipedia QA benchmarks: HotpotQA (HQA)~\citep{yang-etal-2018-hotpotqa}, 2WikiMultihopQA (2Wiki) \citep{ho-etal-2020-constructing}, MuSiQue (MSQ)~\citep{musique}, and Bamboogle (Bamb)~\citep{press2023measuring}.
In addition, we include FRAMES~\citep{krishna-etal-2025-fact} for a challenging multi-hop evaluation.
We use Exact Match (EM) accuracy as the evaluation metric for all datasets.

\paragraph{Search Agent Baselines.} A majority of existing search agent baselines rely on Qwen2.5 variants as the base model~\citep{qwen25}. As prior work does not provide sufficient details on the training setup, data or code, we are unable to reproduce them with Qwen3-4B. 
Therefore, to enable a fair comparison, we gather numbers from InfoSeek~\citep{infoseek}, with an identical setup as ours, in terms of the corpus, search tool model and configuration. 
Next, we demonstrate the effectiveness of the \orbit dataset, by training two comparable search agent baselines using Qwen3-4B as the base model~\citep{qwen3}: 

\begin{enumerate}[leftmargin=*, topsep=0.5pt]
    \item \textbf{Search-R1-4B}: a reproduction of Search-R1~\citep{search_r1}, trained using GRPO with the Qwen3-4B base model on the complete NQ and HotpotQA dataset. 
    \item \textbf{InfoSeeker-4B}: a reproduction of InfoSeeker~\citep{infoseek}, trained using GRPO with the Qwen3-4B base model on an equal mixture (1:~1:~1) of NQ, HotpotQA and InfoSeek.
\end{enumerate} 


\begin{figure*}[t]
	\centering
	\includegraphics[width=0.98\linewidth]{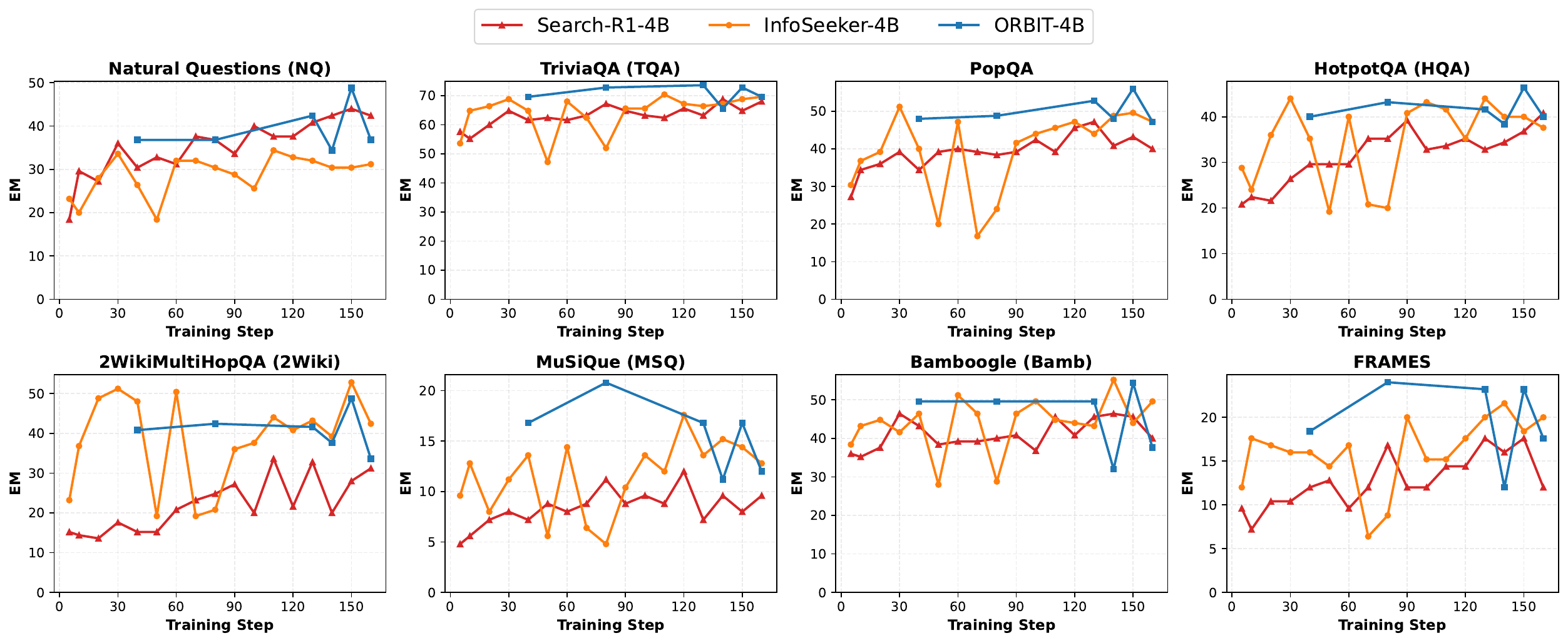}
    \caption{Validation EM accuracy of Search-R1-4B, InfoSeeker-4B and \orbit-4B on 160 training steps on Wikipedia datasets, each with 125 randomly sampled validation pairs. The accuracy drops observed during training primarily occur due to DDGS web search retriever, that can potentially downgrade search results when servers are busy (e.g., \texttt{google} $\rightarrow$ \texttt{bing}). }
	\label{fig_em}
    \vspace{-1.em}
\end{figure*}

\section{Experimental Results \& Ablations}

\paragraph{Overall Performance.} We first discuss the main Wikipedia evaluation results in \autoref{tab:results}. Using Qwen3-4B as the base model, \orbit-4B consistently outperforms Search-R1-4B and InfoSeeker-4B on all single-hop and multi-hop QA Wikipedia datasets, by achieving 49.1 EM accuracy on average, that is 9.0 points better than Search-R1-4B and 4.3 points better than InfoSeeker-4B. This shows the importance of the \orbit dataset over traditional datasets such as NQ and HotpotQA, and similar reasoning-intensive datasets such as InfoSeek. We observe similar gains with \orbit-4B in the Search-R1 evaluation setup in \autoref{app:extended-comparison} and in the validation EM accuracy scores (up to 160 steps) on all datasets provided in \autoref{fig_em}.

\paragraph{Comparison with Qwen2.5-3B-Instruct as the Base Model.} From \autoref{tab:results}, \orbit-4B is compared with several baselines trained using Qwen2.5-3B-Instruct as the base model, taken from InfoSeek~\citep{infoseek}. 
We observe that direct GRPO fine-tuning on a newer and recent base model (e.g., Qwen3-4B), significantly outperforms search agents using Qwen2.5-3B-Instruct as the base model. 
Supervised fine-tuning (SFT) is beneficial for the Qwen2.5 as the base model as an alignment stage. However, GRPO training is sufficient for training Qwen3-4B as the base model. InfoSeeker-3B is the strongest Qwen2.5-3B-Instruct baseline achieving 42.7 EM score on average, 6.4 points below \orbit-4B.

\begin{wraptable}{r}{0.4\columnwidth}
\vspace{-6pt}  
\centering
{\setlength{\tabcolsep}{6pt}
\renewcommand{\arraystretch}{1}
\fontsize{10}{12}\selectfont
\begin{tabular}{@{} l c @{}}
\toprule
\textbf{Model} & \textbf{FRAMES} \\
\midrule
Search-R1-4B$^\dagger$ & 14.7 \\
InfoSeeker-4B$^\dagger$ & 18.5 \\
\rowcolor{odysseyrow} \textcolor{paperblue}{\textbf{\orbit-4B$^\dagger$}} &  \textcolor{paperblue}{\textbf{24.2}} \\
\bottomrule
\end{tabular}
}
\caption{EM accuracy on FRAMES.}
\label{tab:frames_results}
\vspace{-8pt}
\end{wraptable}

\paragraph{Extended Results on FRAMES.} We extend our evaluation on FRAMES~\citep{krishna-etal-2025-fact}, a recent and challenging multi-hop Wikipedia QA dataset. From \autoref{tab:frames_results}, we observe that \orbit-4B achieves the highest EM accuracy in contrast to Search-R1-4B and InfoSeeker-4B.
Although HotpotQA, NQ and InfoSeek are all in-domain, i.e., Wikipedia datasets, fine-tuning on \orbit (Wikipedia and Web) helps in answering multi-hop reasoning questions in FRAMES.

\paragraph{Dataset Mixing Ratios.} We inspect how dataset composition affects the EM accuracy of \orbit-4B. Specifically, we vary the distribution of the training pairs from three datasets, NQ: HotpotQA: \orbit at (1:~2:~4) and our default (1:~1:~1). The (1:~2:~4) dataset mixing ratio reduces the distribution of the single-hop pairs and provides more focus on multi-hop training pairs. 
These comparative results are detailed in \autoref{tab:ablation-mix}. The optimal setting is the equal dataset mix (1:~1:~1), achieving 46.0 EM accuracy, on average. This suggests that for smaller search agents, training directly on challenging multi-hop questions isn't the optimal strategy. 
Rather training on an equally mixed training dataset with easy single-hop or multi-hop questions present in Natural Questions and HotpotQA, helps to teach search agents effectively on answering questions requiring multi-hop steps of information.

\begin{table}[t]
\centering
\renewcommand{\arraystretch}{1.2}
\resizebox{\textwidth}{!}{
\begin{tabular}{lcccccccccc}
\toprule
\multirow{2}{*}{\textbf{\orbit-4B}} & \multicolumn{3}{c}{\textbf{Single-Hop QA}} & \multicolumn{5}{c}{\textbf{Multi-Hop QA}} & \multirow{2}{*}{\textbf{Avg. 8}}\\
\cmidrule(lr){2-4} \cmidrule(lr){5-9}
 & NQ & TQA & PopQA & HQA & 2Wiki & MSQ & Bamb & FRAMES &  \\
\midrule
\rowcolor{odysseyrow} \multicolumn{10}{l}{\textbf{Dataset Mixing Ratios} (Dataset ratio of NQ:~HotpotQA:~\orbit)} \\
\textbf{Ratio (1:~1:~1)}  &
\textcolor{paperblue}{\textbf{43.7}} & \textcolor{paperblue}{\textbf{67.3}} & \textcolor{paperblue}{\textbf{53.8}} & \textcolor{paperblue}{\textbf{42.5}} & 61.1 & \textcolor{paperblue}{\textbf{20.1}} & \textcolor{paperblue}{\textbf{55.2}} & \textcolor{paperblue}{\textbf{24.2}} & \textcolor{paperblue}{\textbf{46.0}} \\
\textbf{Ratio (1:~2:~4)} &
41.7 & 65.3 & 50.8 & 40.0 & \textcolor{paperblue}{\textbf{61.3}} & 16.1 & 52.0 & 19.1 & 43.3 \\
\midrule
\rowcolor{odysseyrow} \multicolumn{10}{l}{\textbf{Top \#K Search Results} (Snippet \& title only during training)} \\
\textbf{Top 5}  &
43.7 & \textcolor{paperblue}{\textbf{67.3}} & 53.8 & 42.5 & \textcolor{paperblue}{\textbf{61.1}} & 20.1 & \textcolor{paperblue}{\textbf{55.2}} & \textcolor{paperblue}{\textbf{24.2}} & \textcolor{paperblue}{\textbf{46.0}} \\
\textbf{Top 10} &
\textcolor{paperblue}{\textbf{43.9}} & 66.0 & \textcolor{paperblue}{\textbf{55.2}} & \textcolor{paperblue}{\textbf{43.1}} & 58.4 & \textcolor{paperblue}{\textbf{21.0}} & 51.2 & 21.4 & 45.0 \\
\bottomrule
\end{tabular}}
\caption{Ablations on dataset mixing ratios and top-$k$ search results with \orbit-4B.}
\vspace{-3mm}
\label{tab:ablation-mix}
\end{table}

\paragraph{Top-K Search Results.} Further, we also inspect how the top-k DDGS search results during each turn affect the EM accuracy of \orbit-4B. 
As shown in \autoref{tab:ablation-mix}, the results are counterintuitive, as more search results do not correlate to better EM accuracy. 
In \orbit-4B, retrieving top 5 search results offers a better trade-off in terms of search latency. We suspect that the search agent observes relevant titles and snippets by avoiding additional search distractors, which lowers the EM accuracy marginally by 1.0 point, on average. 

\section{Discussion}

\paragraph{Dataset Generation Under Budget Constraints.} An important objective is the ability to fully open-source and generate a synthetic training dataset under a tight budget. However, the applicability of the framework is not restricted to a frugal setup. With larger budgets, the framework can be constructed more reliably with paid access to capable LLMs and search APIs, reducing latency to construct faster and better quality synthetic training datasets. In this work, we demonstrate that even with minimal cost and a single laptop, a competitive and reasoning-intensive synthetic training dataset for search agents can be generated.

\paragraph{Training Setup Limitations.}
Due to budget constraints, we focus on small search agents and a relatively minimal training setup. 
Our GRPO training setup is sufficient to demonstrate the effectiveness of \orbit over existing baselines with comparable resources. 
However, to close the gap with larger open-source and proprietary search agents, several extensions would be necessary: (1) access to paid search APIs; (2) an additional \texttt{get\_document} tool that fetches and summarizes the full content of a retrieved webpage; and (3) a larger base model ($\geq$ 30B parameters) combined with SFT warm-up followed by RL fine-tuning~\citep{infoseek, inforage}. We leave these directions as future work.

\section{Conclusion}

We introduce \orbit, a scalable and verifiable dataset with 20K reasoning-intensive question--answer pairs for training search agents on a tight budget. The dataset is built with a four-stage framework that combines seed construction, question--answer pair generation, self-verification, and external verification, without relying on paid search APIs or expensive human annotation. 
Using \orbit, we train \orbit-4B, a search agent using Qwen3-4B as the base model optimized with GRPO. \orbit-4B achieves strong performance among sub-4B search agents and consistently outperforms the baselines on single-hop and multi-hop Wikipedia QA datasets. 
Overall, our findings demonstrate that careful synthetic data generation is a viable path for improving search agents without proprietary tools or large annotation budgets. 
We hope \orbit lowers the barrier to research on deep search and motivates future work on open, reproducible data pipelines for open-source search agents.

\section*{Acknowledgments}
We begin by acknowledging the old but resilient Nandan's 2018 Linux laptop used to construct the \orbit dataset. We sincerely thank DeepSeek for providing an accessible chat interface with integrated web search capabilities. We are also grateful to the Digital Research Alliance of Canada for providing access to the Nibi \& Fir clusters (only clusters with internet access) used for training search agents with 4xH100's. This research was supported in part by the Natural Sciences and Engineering Research Council (NSERC) of Canada.

\section*{Ethics Statement}

\orbit is constructed from publicly accessible sources, primarily Wikipedia categories, publicly retrievable and sourced webpages, and generated question--answer pairs. 
We do not intentionally collect private, sensitive, or personally identifying information, and the dataset is intended for factual question answering research rather than profiling individuals. 

Stages 2 and 3 of the \orbit framework rely on automated interactions with Python Selenium using the DeepSeek Chat web interface (\texttt{chat.deepseek.com}). 
To ensure an ethical practice, we took the following steps: First, all interactions with DeepSeek Chat are performed using the publicly accessible web interface under a \emph{single authenticated personal account}, consistent with how a user would use the chat service manually. 
Second, the automation is \emph{rate-limited} and kept sequential, where each prompt is submitted \emph{one at a time} and the Selenium script waits for a complete model response before proceeding for the next one. 
Third, the purpose of the dataset construction is \emph{purely academic research} with no commercial intent, and the generated content of the dataset will be licensed under the CC BY-NC-SA, to avoid commercial usage. 
Fourth, the study follows the DeepSeek Terms of Use\footnote{\url{https://cdn.deepseek.com/policies/en-US/deepseek-terms-of-use.html}}, where it states that inputs and outputs from the DeepSeek Chat service \emph{can be applied towards academic research}. 
We will explicitly highlight these artifacts with our dataset release; researchers who adopt our dataset should review the applicable terms of any service they use and seek guidance where appropriate.

\section*{Reproducibility Statement}

We aim to make the dataset construction process as reproducible as possible. 
Due to budget constraints, we use the free and open-source DDGS search aggregator that reduces search reproducibility in our work, as the search results can vary based on the DDGS server load. 
Similarly, at the time of \orbit dataset construction, we interacted with \texttt{DeepSeek-V3.1} via DeepSeek Chat; however, the models continue to upgrade and the exact model might not be available at a future date, leading to change in outputs. 
To mitigate this, we are fully open-sourcing and releasing intermediate artifacts wherever possible, including question--answer pairs, cited URLs, verification metadata and experiment configurations, to enable other researchers that can easily construct synthetic training data with limited access to budget, making it a lot more accessible to a wider community.

\bibliography{colm2026}

\begin{thebibliography}{55}
\providecommand{\natexlab}[1]{#1}
\providecommand{\url}[1]{\texttt{#1}}
\expandafter\ifx\csname urlstyle\endcsname\relax
  \providecommand{\doi}[1]{doi: #1}\else
  \providecommand{\doi}{doi: \begingroup \urlstyle{rm}\Url}\fi

\bibitem[Barbaresi(2021)]{barbaresi-2021-trafilatura}
Adrien Barbaresi.
\newblock Trafilatura: {A} web scraping library and command-line tool for text discovery and extraction.
\newblock In Heng Ji, Jong~C. Park, and Rui Xia (eds.), \emph{Proceedings of the 59th Annual Meeting of the Association for Computational Linguistics and the 11th International Joint Conference on Natural Language Processing: System Demonstrations}, pp.\  122--131, Online, August 2021. Association for Computational Linguistics.
\newblock \doi{10.18653/v1/2021.acl-demo.15}.
\newblock URL \url{https://aclanthology.org/2021.acl-demo.15/}.

\bibitem[Chen et~al.(2024)Chen, Xiao, Zhang, Luo, Lian, and Liu]{bge-m3}
Jianlyu Chen, Shitao Xiao, Peitian Zhang, Kun Luo, Defu Lian, and Zheng Liu.
\newblock {M}3-embedding: Multi-linguality, multi-functionality, multi-granularity text embeddings through self-knowledge distillation.
\newblock In Lun-Wei Ku, Andre Martins, and Vivek Srikumar (eds.), \emph{Findings of the Association for Computational Linguistics: ACL 2024}, pp.\  2318--2335, Bangkok, Thailand, August 2024. Association for Computational Linguistics.
\newblock \doi{10.18653/v1/2024.findings-acl.137}.
\newblock URL \url{https://aclanthology.org/2024.findings-acl.137/}.

\bibitem[Chen et~al.(2025)Chen, Sun, Li, sunhaoze, ZhouYijie, Zhu, Wang, Pan, Zhang, Chen, Yang, Zhou, and Chen]{reSearch}
Mingyang Chen, Linzhuang Sun, Tianpeng Li, sunhaoze, ZhouYijie, Chenzheng Zhu, Haofen Wang, Jeff~Z. Pan, Wen Zhang, Huajun Chen, Fan Yang, Zenan Zhou, and Weipeng Chen.
\newblock {ReSearch}: Learning to reason with search for {LLM}s via reinforcement learning.
\newblock In \emph{The Thirty-ninth Annual Conference on Neural Information Processing Systems}, 2025.
\newblock URL \url{https://openreview.net/forum?id=OuGAwwAT8G}.

\bibitem[Gao et~al.(2023)Gao, Yen, Yu, and Chen]{gao-etal-2023-enabling}
Tianyu Gao, Howard Yen, Jiatong Yu, and Danqi Chen.
\newblock Enabling large language models to generate text with citations.
\newblock In Houda Bouamor, Juan Pino, and Kalika Bali (eds.), \emph{Proceedings of the 2023 Conference on Empirical Methods in Natural Language Processing}, pp.\  6465--6488, Singapore, December 2023. Association for Computational Linguistics.
\newblock \doi{10.18653/v1/2023.emnlp-main.398}.
\newblock URL \url{https://aclanthology.org/2023.emnlp-main.398/}.

\bibitem[Gemini(2025)]{gemini2025deepresearch}
Gemini.
\newblock {Gemini Deep Research}.
\newblock 2025.
\newblock URL \url{https://gemini.google/overview/deep-research/}.

\bibitem[Ho et~al.(2020{\natexlab{a}})Ho, Duong~Nguyen, Sugawara, and Aizawa]{2wikimultihopqa}
Xanh Ho, Anh-Khoa Duong~Nguyen, Saku Sugawara, and Akiko Aizawa.
\newblock Constructing a multi-hop {QA} dataset for comprehensive evaluation of reasoning steps.
\newblock In Donia Scott, Nuria Bel, and Chengqing Zong (eds.), \emph{Proceedings of the 28th International Conference on Computational Linguistics}, pp.\  6609--6625, Barcelona, Spain (Online), December 2020{\natexlab{a}}. International Committee on Computational Linguistics.
\newblock \doi{10.18653/v1/2020.coling-main.580}.
\newblock URL \url{https://aclanthology.org/2020.coling-main.580/}.

\bibitem[Ho et~al.(2020{\natexlab{b}})Ho, Duong~Nguyen, Sugawara, and Aizawa]{ho-etal-2020-constructing}
Xanh Ho, Anh-Khoa Duong~Nguyen, Saku Sugawara, and Akiko Aizawa.
\newblock Constructing a multi-hop {QA} dataset for comprehensive evaluation of reasoning steps.
\newblock In Donia Scott, Nuria Bel, and Chengqing Zong (eds.), \emph{Proceedings of the 28th International Conference on Computational Linguistics}, pp.\  6609--6625, Barcelona, Spain (Online), December 2020{\natexlab{b}}. International Committee on Computational Linguistics.
\newblock \doi{10.18653/v1/2020.coling-main.580}.
\newblock URL \url{https://aclanthology.org/2020.coling-main.580/}.

\bibitem[Jiang et~al.(2025)Jiang, Lu, Li, Lyu, Nie, Wang, Su, Chen, Zou, Du, Pang, and Chen]{2025verltool}
Dongfu Jiang, Yi~Lu, Zhuofeng Li, Zhiheng Lyu, Ping Nie, Haozhe Wang, Alex Su, Hui Chen, Kai Zou, Chao Du, Tianyu Pang, and Wenhu Chen.
\newblock {VerlTool: Towards Holistic Agentic Reinforcement Learning with Tool Use}, 2025.
\newblock URL \url{https://arxiv.org/abs/2509.01055}.

\bibitem[Jiang et~al.(2023)Jiang, Xu, Gao, Sun, Liu, Dwivedi-Yu, Yang, Callan, and Neubig]{jiang-etal-2023-active}
Zhengbao Jiang, Frank Xu, Luyu Gao, Zhiqing Sun, Qian Liu, Jane Dwivedi-Yu, Yiming Yang, Jamie Callan, and Graham Neubig.
\newblock Active retrieval augmented generation.
\newblock In Houda Bouamor, Juan Pino, and Kalika Bali (eds.), \emph{Proceedings of the 2023 Conference on Empirical Methods in Natural Language Processing}, pp.\  7969--7992, Singapore, December 2023. Association for Computational Linguistics.
\newblock \doi{10.18653/v1/2023.emnlp-main.495}.
\newblock URL \url{https://aclanthology.org/2023.emnlp-main.495/}.

\bibitem[Jin et~al.(2025)Jin, Zeng, Yue, Yoon, Arik, Wang, Zamani, and Han]{search_r1}
Bowen Jin, Hansi Zeng, Zhenrui Yue, Jinsung Yoon, Sercan~O Arik, Dong Wang, Hamed Zamani, and Jiawei Han.
\newblock {Search-R1}: Training {LLM}s to reason and leverage search engines with reinforcement learning.
\newblock In \emph{Second Conference on Language Modeling}, 2025.
\newblock URL \url{https://openreview.net/forum?id=Rwhi91ideu}.

\bibitem[Joshi et~al.(2017)Joshi, Choi, Weld, and Zettlemoyer]{joshi-etal-2017-triviaqa}
Mandar Joshi, Eunsol Choi, Daniel Weld, and Luke Zettlemoyer.
\newblock {T}rivia{QA}: A large scale distantly supervised challenge dataset for reading comprehension.
\newblock In Regina Barzilay and Min-Yen Kan (eds.), \emph{Proceedings of the 55th Annual Meeting of the Association for Computational Linguistics (Volume 1: Long Papers)}, pp.\  1601--1611, Vancouver, Canada, July 2017. Association for Computational Linguistics.
\newblock \doi{10.18653/v1/P17-1147}.
\newblock URL \url{https://aclanthology.org/P17-1147/}.

\bibitem[Karpukhin et~al.(2020)Karpukhin, Oguz, Min, Lewis, Wu, Edunov, Chen, and Yih]{karpukhin-etal-2020-dense}
Vladimir Karpukhin, Barlas Oguz, Sewon Min, Patrick Lewis, Ledell Wu, Sergey Edunov, Danqi Chen, and Wen-tau Yih.
\newblock Dense passage retrieval for open-domain question answering.
\newblock In Bonnie Webber, Trevor Cohn, Yulan He, and Yang Liu (eds.), \emph{Proceedings of the 2020 Conference on Empirical Methods in Natural Language Processing (EMNLP)}, pp.\  6769--6781, Online, November 2020. Association for Computational Linguistics.
\newblock \doi{10.18653/v1/2020.emnlp-main.550}.
\newblock URL \url{https://aclanthology.org/2020.emnlp-main.550/}.

\bibitem[Khandelwal et~al.(2020)Khandelwal, Levy, Jurafsky, Zettlemoyer, and Lewis]{khandelwal:2020}
Urvashi Khandelwal, Omer Levy, Dan Jurafsky, Luke Zettlemoyer, and Mike Lewis.
\newblock Generalization through memorization: Nearest neighbor language models.
\newblock In \emph{International Conference on Learning Representations}, 2020.
\newblock URL \url{https://openreview.net/forum?id=HklBjCEKvH}.

\bibitem[Krishna et~al.(2025)Krishna, Krishna, Mohananey, Schwarcz, Stambler, Upadhyay, and Faruqui]{krishna-etal-2025-fact}
Satyapriya Krishna, Kalpesh Krishna, Anhad Mohananey, Steven Schwarcz, Adam Stambler, Shyam Upadhyay, and Manaal Faruqui.
\newblock Fact, fetch, and reason: A unified evaluation of retrieval-augmented generation.
\newblock In Luis Chiruzzo, Alan Ritter, and Lu~Wang (eds.), \emph{Proceedings of the 2025 Conference of the Nations of the Americas Chapter of the Association for Computational Linguistics: Human Language Technologies (Volume 1: Long Papers)}, pp.\  4745--4759, Albuquerque, New Mexico, April 2025. Association for Computational Linguistics.
\newblock ISBN 979-8-89176-189-6.
\newblock \doi{10.18653/v1/2025.naacl-long.243}.
\newblock URL \url{https://aclanthology.org/2025.naacl-long.243/}.

\bibitem[Kwiatkowski et~al.(2019)Kwiatkowski, Palomaki, Redfield, Collins, Parikh, Alberti, Epstein, Polosukhin, Devlin, Lee, Toutanova, Jones, Kelcey, Chang, Dai, Uszkoreit, Le, and Petrov]{kwiatkowski-etal-2019-natural}
Tom Kwiatkowski, Jennimaria Palomaki, Olivia Redfield, Michael Collins, Ankur Parikh, Chris Alberti, Danielle Epstein, Illia Polosukhin, Jacob Devlin, Kenton Lee, Kristina Toutanova, Llion Jones, Matthew Kelcey, Ming-Wei Chang, Andrew~M. Dai, Jakob Uszkoreit, Quoc Le, and Slav Petrov.
\newblock Natural questions: A benchmark for question answering research.
\newblock \emph{Transactions of the Association for Computational Linguistics}, 7:\penalty0 452--466, 2019.
\newblock \doi{10.1162/tacl_a_00276}.
\newblock URL \url{https://aclanthology.org/Q19-1026/}.

\bibitem[Lewis et~al.(2020)Lewis, Perez, Piktus, Petroni, Karpukhin, Goyal, K\"{u}ttler, Lewis, Yih, Rockt\"{a}schel, Riedel, and Kiela]{lewis:2020}
Patrick Lewis, Ethan Perez, Aleksandra Piktus, Fabio Petroni, Vladimir Karpukhin, Naman Goyal, Heinrich K\"{u}ttler, Mike Lewis, Wen-tau Yih, Tim Rockt\"{a}schel, Sebastian Riedel, and Douwe Kiela.
\newblock Retrieval-augmented generation for knowledge-intensive nlp tasks.
\newblock In H.~Larochelle, M.~Ranzato, R.~Hadsell, M.F. Balcan, and H.~Lin (eds.), \emph{Advances in Neural Information Processing Systems}, volume~33, pp.\  9459--9474. Curran Associates, Inc., 2020.
\newblock URL \url{https://proceedings.neurips.cc/paper_files/paper/2020/file/6b493230205f780e1bc26945df7481e5-Paper.pdf}.

\bibitem[Li et~al.(2026{\natexlab{a}})Li, Chen, Xu, Jin, Wu, Wang, Yuan, and Wang]{li2026improvingsearchagentline}
Jian Li, Dongsheng Chen, Zhenhua Xu, Yizhang Jin, Jiafu Wu, Chengjie Wang, Xiaotong Yuan, and Yabiao Wang.
\newblock Improving search agent with one line of code, 2026{\natexlab{a}}.
\newblock URL \url{https://arxiv.org/abs/2603.10069}.

\bibitem[Li et~al.(2026{\natexlab{b}})Li, Jin, Liu, Ding, Wu, Chen, Shen, Qin, Tai, Wang, Yuan, and Wang]{li2026sesearchselfevolvingsearchagent}
Jian Li, Yizhang Jin, Dongqi Liu, Hang Ding, Jiafu Wu, Dongsheng Chen, Yunhang Shen, Yulei Qin, Ying Tai, Chengjie Wang, Xiaotong Yuan, and Yabiao Wang.
\newblock {SE-Search}: Self-evolving search agent via memory and dense reward, 2026{\natexlab{b}}.
\newblock URL \url{https://arxiv.org/abs/2603.03293}.

\bibitem[Li et~al.(2025{\natexlab{a}})Li, Zhang, Yin, Zhang, Ou, Wu, Yin, Li, Tao, Wang, Shen, Zhang, Zhang, Wu, Jiang, Yan, Xie, Huang, and Zhou]{websailor}
Kuan Li, Zhongwang Zhang, Huifeng Yin, Liwen Zhang, Litu Ou, Jialong Wu, Wenbiao Yin, Baixuan Li, Zhengwei Tao, Xinyu Wang, Weizhou Shen, Junkai Zhang, Dingchu Zhang, Xixi Wu, Yong Jiang, Ming Yan, Pengjun Xie, Fei Huang, and Jingren Zhou.
\newblock {WebSailor: Navigating Super-human Reasoning for Web Agent}, 2025{\natexlab{a}}.
\newblock URL \url{https://arxiv.org/abs/2507.02592}.

\bibitem[Li et~al.(2025{\natexlab{b}})Li, Dong, Jin, Zhang, Zhou, Zhu, Zhang, and Dou]{search_o1}
Xiaoxi Li, Guanting Dong, Jiajie Jin, Yuyao Zhang, Yujia Zhou, Yutao Zhu, Peitian Zhang, and Zhicheng Dou.
\newblock Search-o1: Agentic search-enhanced large reasoning models.
\newblock In Christos Christodoulopoulos, Tanmoy Chakraborty, Carolyn Rose, and Violet Peng (eds.), \emph{Proceedings of the 2025 Conference on Empirical Methods in Natural Language Processing}, pp.\  5420--5438, Suzhou, China, November 2025{\natexlab{b}}. Association for Computational Linguistics.
\newblock ISBN 979-8-89176-332-6.
\newblock \doi{10.18653/v1/2025.emnlp-main.276}.
\newblock URL \url{https://aclanthology.org/2025.emnlp-main.276/}.

\bibitem[Liu et~al.(2025)Liu, Li, Zhang, Li, Chen, Ji, Cheng, Wu, Du, Xu, Song, Zhu, Chen, Zhao, and He]{webexplorer}
Junteng Liu, Yunji Li, Chi Zhang, Jingyang Li, Aili Chen, Ke~Ji, Weiyu Cheng, Zijia Wu, Chengyu Du, Qidi Xu, Jiayuan Song, Zhengmao Zhu, Wenhu Chen, Pengyu Zhao, and Junxian He.
\newblock {WebExplorer: Explore and Evolve for Training Long-Horizon Web Agents}, 2025.
\newblock URL \url{https://arxiv.org/abs/2509.06501}.

\bibitem[Lu et~al.(2025)Lu, Hou, Wang, Zhang, Liu, Li, Feng, Tang, and Dong]{deepdive}
Rui Lu, Zhenyu Hou, Zihan Wang, Hanchen Zhang, Xiao Liu, Yujiang Li, Shi Feng, Jie Tang, and Yuxiao Dong.
\newblock {DeepDive: Advancing Deep Search Agents with Knowledge Graphs and Multi-Turn RL}, 2025.
\newblock URL \url{https://arxiv.org/abs/2509.10446}.

\bibitem[Mallen et~al.(2023)Mallen, Asai, Zhong, Das, Khashabi, and Hajishirzi]{mallen-etal-2023-trust}
Alex Mallen, Akari Asai, Victor Zhong, Rajarshi Das, Daniel Khashabi, and Hannaneh Hajishirzi.
\newblock When not to trust language models: Investigating effectiveness of parametric and non-parametric memories.
\newblock In Anna Rogers, Jordan Boyd-Graber, and Naoaki Okazaki (eds.), \emph{Proceedings of the 61st Annual Meeting of the Association for Computational Linguistics (Volume 1: Long Papers)}, pp.\  9802--9822, Toronto, Canada, July 2023. Association for Computational Linguistics.
\newblock \doi{10.18653/v1/2023.acl-long.546}.
\newblock URL \url{https://aclanthology.org/2023.acl-long.546/}.

\bibitem[Mei et~al.(2025)Mei, Hu, Fu, Wen, Yang, Wu, Cai, Cai, Gao, Yang, Xie, Shi, Liu, and Qiao]{mei:2025}
Jianbiao Mei, Tao Hu, Daocheng Fu, Licheng Wen, Xuemeng Yang, Rong Wu, Pinlong Cai, Xinyu Cai, Xing Gao, Yu~Yang, Chengjun Xie, Botian Shi, Yong Liu, and Yu~Qiao.
\newblock O$^2$-searcher: A searching-based agent model for open-domain open-ended question answering, 2025.
\newblock URL \url{https://arxiv.org/abs/2505.16582}.

\bibitem[OpenAI(2025)]{openai2025deepresearch}
OpenAI.
\newblock {Introducing Deep Research}.
\newblock 2025.
\newblock URL \url{https://openai.com/index/introducing-deep-research/}.

\bibitem[Panickssery et~al.(2024)Panickssery, Bowman, and Feng]{panickssery2024llm}
Arjun Panickssery, Samuel~R. Bowman, and Shi Feng.
\newblock {LLM} evaluators recognize and favor their own generations.
\newblock In \emph{The Thirty-eighth Annual Conference on Neural Information Processing Systems}, 2024.
\newblock URL \url{https://openreview.net/forum?id=4NJBV6Wp0h}.

\bibitem[Petroni et~al.(2021)Petroni, Piktus, Fan, Lewis, Yazdani, De~Cao, Thorne, Jernite, Karpukhin, Maillard, Plachouras, Rockt{\"a}schel, and Riedel]{petroni-etal-2021-kilt}
Fabio Petroni, Aleksandra Piktus, Angela Fan, Patrick Lewis, Majid Yazdani, Nicola De~Cao, James Thorne, Yacine Jernite, Vladimir Karpukhin, Jean Maillard, Vassilis Plachouras, Tim Rockt{\"a}schel, and Sebastian Riedel.
\newblock {KILT}: a benchmark for knowledge intensive language tasks.
\newblock In Kristina Toutanova, Anna Rumshisky, Luke Zettlemoyer, Dilek Hakkani-Tur, Iz~Beltagy, Steven Bethard, Ryan Cotterell, Tanmoy Chakraborty, and Yichao Zhou (eds.), \emph{Proceedings of the 2021 Conference of the North American Chapter of the Association for Computational Linguistics: Human Language Technologies}, pp.\  2523--2544, Online, June 2021. Association for Computational Linguistics.
\newblock \doi{10.18653/v1/2021.naacl-main.200}.
\newblock URL \url{https://aclanthology.org/2021.naacl-main.200/}.

\bibitem[Press et~al.(2023)Press, Zhang, Min, Schmidt, Smith, and Lewis]{press2023measuring}
Ofir Press, Muru Zhang, Sewon Min, Ludwig Schmidt, Noah~A. Smith, and Mike Lewis.
\newblock Measuring and narrowing the compositionality gap in language models, 2023.
\newblock URL \url{https://openreview.net/forum?id=PUwbwZJz9dO}.

\bibitem[Qian \& Liu(2025)Qian and Liu]{inforage}
Hongjin Qian and Zheng Liu.
\newblock Scent of knowledge: Optimizing search-enhanced reasoning with information foraging.
\newblock In \emph{The Thirty-ninth Annual Conference on Neural Information Processing Systems}, 2025.
\newblock URL \url{https://openreview.net/forum?id=26kUrQm4zw}.

\bibitem[Shao et~al.(2024)Shao, Wang, Zhu, Xu, Song, Bi, Zhang, Zhang, Li, Wu, and Guo]{shao:2024}
Zhihong Shao, Peiyi Wang, Qihao Zhu, Runxin Xu, Junxiao Song, Xiao Bi, Haowei Zhang, Mingchuan Zhang, Y.~K. Li, Y.~Wu, and Daya Guo.
\newblock {DeepSeekMath: Pushing the Limits of Mathematical Reasoning in Open Language Models}, 2024.
\newblock URL \url{https://arxiv.org/abs/2402.03300}.

\bibitem[Shi et~al.(2025{\natexlab{a}})Shi, Li, Wu, Liu, Fang, Cai, Zhang, and Wang]{shi2025search}
Yaorui Shi, Sihang Li, Chang Wu, Zhiyuan Liu, Junfeng Fang, Hengxing Cai, An~Zhang, and Xiang Wang.
\newblock Search and refine during think: Facilitating knowledge refinement for improved retrieval-augmented reasoning.
\newblock In \emph{The Thirty-ninth Annual Conference on Neural Information Processing Systems}, 2025{\natexlab{a}}.
\newblock URL \url{https://openreview.net/forum?id=rBlWKIUQey}.

\bibitem[Shi et~al.(2025{\natexlab{b}})Shi, Yan, Yin, Verberne, de~Rijke, and Ren]{shi2025iterative}
Zhengliang Shi, Lingyong Yan, Dawei Yin, Suzan Verberne, Maarten de~Rijke, and Zhaochun Ren.
\newblock Iterative self-incentivization empowers large language models as agentic searchers.
\newblock In \emph{The Thirty-ninth Annual Conference on Neural Information Processing Systems}, 2025{\natexlab{b}}.
\newblock URL \url{https://openreview.net/forum?id=s9NkfkUuEr}.

\bibitem[Song et~al.(2025)Song, Jiang, Min, Chen, Chen, Zhao, Fang, and Wen]{r1_searcher}
Huatong Song, Jinhao Jiang, Yingqian Min, Jie Chen, Zhipeng Chen, Wayne~Xin Zhao, Lei Fang, and Ji-Rong Wen.
\newblock {R1-Searcher: Incentivizing the Search Capability in LLMs via Reinforcement Learning}, 2025.
\newblock URL \url{https://arxiv.org/abs/2503.05592}.

\bibitem[Sun et~al.(2025{\natexlab{a}})Sun, Qiao, Guo, Fan, Hou, Jiang, Xie, Zhang, Huang, and Zhou]{zerosearchbase:2025}
Hao Sun, Zile Qiao, Jiayan Guo, Xuanbo Fan, Yingyan Hou, Yong Jiang, Pengjun Xie, Yan Zhang, Fei Huang, and Jingren Zhou.
\newblock {ZeroSearch: Incentivize the Search Capability of LLMs without Searching}, 2025{\natexlab{a}}.
\newblock URL \url{https://arxiv.org/abs/2505.04588}.

\bibitem[Sun et~al.(2025{\natexlab{b}})Sun, Song, Wang, Ren, Jiang, Zhang, Bai, Deng, Zhao, Liu, Fang, Wang, and Wen]{simpledeepsearcher}
Shuang Sun, Huatong Song, Yuhao Wang, Ruiyang Ren, Jinhao Jiang, Junjie Zhang, Fei Bai, Jia Deng, Wayne~Xin Zhao, Zheng Liu, Lei Fang, Zhongyuan Wang, and Ji-Rong Wen.
\newblock {S}imple{D}eep{S}earcher: Deep information seeking via web-powered reasoning trajectory synthesis.
\newblock In Christos Christodoulopoulos, Tanmoy Chakraborty, Carolyn Rose, and Violet Peng (eds.), \emph{Findings of the Association for Computational Linguistics: EMNLP 2025}, pp.\  13705--13720, Suzhou, China, November 2025{\natexlab{b}}. Association for Computational Linguistics.
\newblock ISBN 979-8-89176-335-7.
\newblock \doi{10.18653/v1/2025.findings-emnlp.739}.
\newblock URL \url{https://aclanthology.org/2025.findings-emnlp.739/}.

\bibitem[Sutton \& Barto(1998)Sutton and Barto]{sutton:1998}
R.S. Sutton and A.G. Barto.
\newblock Reinforcement learning: An introduction.
\newblock \emph{IEEE Transactions on Neural Networks}, 9\penalty0 (5):\penalty0 1054--1054, Sep. 1998.
\newblock ISSN 1941-0093.
\newblock \doi{10.1109/TNN.1998.712192}.

\bibitem[Tao et~al.(2026)Tao, Wu, Yin, Wu, Zhang, Li, SHEN, Li, Zhang, Wang, Zhang, Jiang, Xie, Huang, and Zhou]{webshaper}
Zhengwei Tao, Jialong Wu, Wenbiao Yin, Pu~Wu, Junkai Zhang, Baixuan Li, Haiyang SHEN, Kuan Li, Liwen Zhang, Xinyu Wang, Wentao Zhang, Yong Jiang, Pengjun Xie, Fei Huang, and Jingren Zhou.
\newblock {WebShaper: Agentically Data Synthesizing via Information-Seeking Formalization}.
\newblock In \emph{The Fourteenth International Conference on Learning Representations}, 2026.
\newblock URL \url{https://openreview.net/forum?id=hld4TzJsnD}.

\bibitem[Trivedi et~al.(2022)Trivedi, Balasubramanian, Khot, and Sabharwal]{musique}
Harsh Trivedi, Niranjan Balasubramanian, Tushar Khot, and Ashish Sabharwal.
\newblock {MuSiQue}: Multihop questions via single-hop question composition.
\newblock \emph{Transactions of the Association for Computational Linguistics}, 10:\penalty0 539--554, 05 2022.
\newblock ISSN 2307-387X.
\newblock \doi{10.1162/tacl_a_00475}.
\newblock URL \url{https://doi.org/10.1162/tacl_a_00475}.

\bibitem[Trivedi et~al.(2023)Trivedi, Balasubramanian, Khot, and Sabharwal]{trivedi-etal-2023-interleaving}
Harsh Trivedi, Niranjan Balasubramanian, Tushar Khot, and Ashish Sabharwal.
\newblock Interleaving retrieval with chain-of-thought reasoning for knowledge-intensive multi-step questions.
\newblock In Anna Rogers, Jordan Boyd-Graber, and Naoaki Okazaki (eds.), \emph{Proceedings of the 61st Annual Meeting of the Association for Computational Linguistics (Volume 1: Long Papers)}, pp.\  10014--10037, Toronto, Canada, July 2023. Association for Computational Linguistics.
\newblock \doi{10.18653/v1/2023.acl-long.557}.
\newblock URL \url{https://aclanthology.org/2023.acl-long.557/}.

\bibitem[Wang et~al.(2024)Wang, Yang, Huang, Jiao, Yang, Jiang, Majumder, and Wei]{wang:2022}
Liang Wang, Nan Yang, Xiaolong Huang, Binxing Jiao, Linjun Yang, Daxin Jiang, Rangan Majumder, and Furu Wei.
\newblock Text embeddings by weakly-supervised contrastive pre-training, 2024.
\newblock URL \url{https://arxiv.org/abs/2212.03533}.

\bibitem[Wei et~al.(2025)Wei, Sun, Papay, McKinney, Han, Fulford, Chung, Passos, Fedus, and Glaese]{browsecomp:2025}
Jason Wei, Zhiqing Sun, Spencer Papay, Scott McKinney, Jeffrey Han, Isa Fulford, Hyung~Won Chung, Alex~Tachard Passos, William Fedus, and Amelia Glaese.
\newblock {BrowseComp: A Simple Yet Challenging Benchmark for Browsing Agents}, 2025.
\newblock URL \url{https://arxiv.org/abs/2504.12516}.

\bibitem[Wen et~al.(2026)Wen, Liu, Zheng, Ye, Wu, Wang, Xu, Liang, Li, Miao, Bian, and Yang]{wen2026reinforcement}
Xumeng Wen, Zihan Liu, Shun Zheng, Shengyu Ye, Zhirong Wu, Yang Wang, Zhijian Xu, Xiao Liang, Junjie Li, Ziming Miao, Jiang Bian, and Mao Yang.
\newblock Reinforcement learning with verifiable rewards implicitly incentivizes correct reasoning in base {LLM}s.
\newblock In \emph{The Fourteenth International Conference on Learning Representations}, 2026.
\newblock URL \url{https://openreview.net/forum?id=jGbRWwIidy}.

\bibitem[Wolfson et~al.(2026)Wolfson, Trivedi, Geva, Goldberg, Roth, Khot, Sabharwal, and Tsarfaty]{monaco}
Tomer Wolfson, Harsh Trivedi, Mor Geva, Yoav Goldberg, Dan Roth, Tushar Khot, Ashish Sabharwal, and Reut Tsarfaty.
\newblock {M}o{N}a{C}o: More natural and complex questions for reasoning across dozens of documents.
\newblock \emph{Transactions of the Association for Computational Linguistics}, 14:\penalty0 23--46, 2026.
\newblock \doi{10.1162/tacl.a.64}.
\newblock URL \url{https://aclanthology.org/2026.tacl-1.2/}.

\bibitem[Wu et~al.(2025)Wu, Yin, Jiang, Wang, Xi, Fang, Zhang, He, Zhou, Xie, and Huang]{webwalker}
Jialong Wu, Wenbiao Yin, Yong Jiang, Zhenglin Wang, Zekun Xi, Runnan Fang, Linhai Zhang, Yulan He, Deyu Zhou, Pengjun Xie, and Fei Huang.
\newblock {W}eb{W}alker: Benchmarking {LLM}s in web traversal.
\newblock In Wanxiang Che, Joyce Nabende, Ekaterina Shutova, and Mohammad~Taher Pilehvar (eds.), \emph{Proceedings of the 63rd Annual Meeting of the Association for Computational Linguistics (Volume 1: Long Papers)}, pp.\  10290--10305, Vienna, Austria, July 2025. Association for Computational Linguistics.
\newblock ISBN 979-8-89176-251-0.
\newblock \doi{10.18653/v1/2025.acl-long.508}.
\newblock URL \url{https://aclanthology.org/2025.acl-long.508/}.

\bibitem[Xi et~al.(2025)Xi, Lin, Zhu, Xiao, Ou, Liu, Wan, Chen, Liu, Wang, Tang, Zhang, and Yu]{infodeepseek}
Yunjia Xi, Jianghao Lin, Menghui Zhu, Yongzhao Xiao, Zhuoying Ou, Jiaqi Liu, Tong Wan, Bo~Chen, Weiwen Liu, Yasheng Wang, Ruiming Tang, Weinan Zhang, and Yong Yu.
\newblock Infodeepseek: Benchmarking agentic information seeking for retrieval-augmented generation, 2025.
\newblock URL \url{https://arxiv.org/abs/2505.15872}.

\bibitem[Xia et~al.(2026)Xia, Luo, Qian, Bao, and Liu]{infoseek}
Ziyi Xia, Kun Luo, Hongjin Qian, Siqi Bao, and Zheng Liu.
\newblock {Open Data Synthesis For Deep Research}.
\newblock In \emph{The Fourteenth International Conference on Learning Representations}, 2026.
\newblock URL \url{https://openreview.net/forum?id=2c9TjRbAib}.

\bibitem[Yang et~al.(2025{\natexlab{a}})Yang, Li, Yang, Zhang, Hui, Zheng, Yu, Gao, Huang, Lv, Zheng, Liu, Zhou, Huang, Hu, Ge, Wei, Lin, Tang, Yang, Tu, Zhang, Yang, Yang, Zhou, Zhou, Lin, Dang, Bao, Yang, Yu, Deng, Li, Xue, Li, Zhang, Wang, Zhu, Men, Gao, Liu, Luo, Li, Tang, Yin, Ren, Wang, Zhang, Ren, Fan, Su, Zhang, Zhang, Wan, Liu, Wang, Cui, Zhang, Zhou, and Qiu]{qwen3}
An~Yang, Anfeng Li, Baosong Yang, Beichen Zhang, Binyuan Hui, Bo~Zheng, Bowen Yu, Chang Gao, Chengen Huang, Chenxu Lv, Chujie Zheng, Dayiheng Liu, Fan Zhou, Fei Huang, Feng Hu, Hao Ge, Haoran Wei, Huan Lin, Jialong Tang, Jian Yang, Jianhong Tu, Jianwei Zhang, Jianxin Yang, Jiaxi Yang, Jing Zhou, Jingren Zhou, Junyang Lin, Kai Dang, Keqin Bao, Kexin Yang, Le~Yu, Lianghao Deng, Mei Li, Mingfeng Xue, Mingze Li, Pei Zhang, Peng Wang, Qin Zhu, Rui Men, Ruize Gao, Shixuan Liu, Shuang Luo, Tianhao Li, Tianyi Tang, Wenbiao Yin, Xingzhang Ren, Xinyu Wang, Xinyu Zhang, Xuancheng Ren, Yang Fan, Yang Su, Yichang Zhang, Yinger Zhang, Yu~Wan, Yuqiong Liu, Zekun Wang, Zeyu Cui, Zhenru Zhang, Zhipeng Zhou, and Zihan Qiu.
\newblock Qwen3 technical report, 2025{\natexlab{a}}.
\newblock URL \url{https://arxiv.org/abs/2505.09388}.

\bibitem[Yang et~al.(2025{\natexlab{b}})Yang, Yang, Zhang, Hui, Zheng, Yu, Li, Liu, Huang, Wei, Lin, Yang, Tu, Zhang, Yang, Yang, Zhou, Lin, Dang, Lu, Bao, Yang, Yu, Li, Xue, Zhang, Zhu, Men, Lin, Li, Tang, Xia, Ren, Ren, Fan, Su, Zhang, Wan, Liu, Cui, Zhang, and Qiu]{qwen25}
An~Yang, Baosong Yang, Beichen Zhang, Binyuan Hui, Bo~Zheng, Bowen Yu, Chengyuan Li, Dayiheng Liu, Fei Huang, Haoran Wei, Huan Lin, Jian Yang, Jianhong Tu, Jianwei Zhang, Jianxin Yang, Jiaxi Yang, Jingren Zhou, Junyang Lin, Kai Dang, Keming Lu, Keqin Bao, Kexin Yang, Le~Yu, Mei Li, Mingfeng Xue, Pei Zhang, Qin Zhu, Rui Men, Runji Lin, Tianhao Li, Tianyi Tang, Tingyu Xia, Xingzhang Ren, Xuancheng Ren, Yang Fan, Yang Su, Yichang Zhang, Yu~Wan, Yuqiong Liu, Zeyu Cui, Zhenru Zhang, and Zihan Qiu.
\newblock {Qwen2.5 Technical Report}, 2025{\natexlab{b}}.
\newblock URL \url{https://arxiv.org/abs/2412.15115}.

\bibitem[Yang et~al.(2018)Yang, Qi, Zhang, Bengio, Cohen, Salakhutdinov, and Manning]{yang-etal-2018-hotpotqa}
Zhilin Yang, Peng Qi, Saizheng Zhang, Yoshua Bengio, William Cohen, Ruslan Salakhutdinov, and Christopher~D. Manning.
\newblock {H}otpot{QA}: A dataset for diverse, explainable multi-hop question answering.
\newblock In Ellen Riloff, David Chiang, Julia Hockenmaier, and Jun{'}ichi Tsujii (eds.), \emph{Proceedings of the 2018 Conference on Empirical Methods in Natural Language Processing}, pp.\  2369--2380, Brussels, Belgium, October-November 2018. Association for Computational Linguistics.
\newblock \doi{10.18653/v1/D18-1259}.
\newblock URL \url{https://aclanthology.org/D18-1259/}.

\bibitem[Yoran et~al.(2024)Yoran, Wolfson, Ram, and Berant]{yoran2024making}
Ori Yoran, Tomer Wolfson, Ori Ram, and Jonathan Berant.
\newblock Making retrieval-augmented language models robust to irrelevant context.
\newblock In \emph{The Twelfth International Conference on Learning Representations}, 2024.
\newblock URL \url{https://openreview.net/forum?id=ZS4m74kZpH}.

\bibitem[Yu et~al.(2025)Yu, Zhang, Zhu, Yuan, Zuo, YuYue, Dai, Fan, Liu, Liu, Liu, Liu, Lin, Lin, Ma, Sheng, Tong, Zhang, Zhang, Zhang, Zhang, Zhu, Zhu, Chen, Chen, Wang, Yu, Song, Wei, Zhou, Liu, Ma, Zhang, Yan, Wu, and Wang]{yu2025dapo}
Qiying Yu, Zheng Zhang, Ruofei Zhu, Yufeng Yuan, Xiaochen Zuo, YuYue, Weinan Dai, Tiantian Fan, Gaohong Liu, Juncai Liu, LingJun Liu, Xin Liu, Haibin Lin, Zhiqi Lin, Bole Ma, Guangming Sheng, Yuxuan Tong, Chi Zhang, Mofan Zhang, Ru~Zhang, Wang Zhang, Hang Zhu, Jinhua Zhu, Jiaze Chen, Jiangjie Chen, Chengyi Wang, Hongli Yu, Yuxuan Song, Xiangpeng Wei, Hao Zhou, Jingjing Liu, Wei-Ying Ma, Ya-Qin Zhang, Lin Yan, Yonghui Wu, and Mingxuan Wang.
\newblock {DAPO}: An open-source {LLM} reinforcement learning system at scale.
\newblock In \emph{The Thirty-ninth Annual Conference on Neural Information Processing Systems}, 2025.
\newblock URL \url{https://openreview.net/forum?id=2a36EMSSTp}.

\bibitem[Yue et~al.(2026)Yue, Upasani, Yang, Ge, Nie, Mao, Liu, and Wang]{dr-zero:2026}
Zhenrui Yue, Kartikeya Upasani, Xianjun Yang, Suyu Ge, Shaoliang Nie, Yuning Mao, Zhe Liu, and Dong Wang.
\newblock {Dr. Zero: Self-Evolving Search Agents without Training Data}, 2026.
\newblock URL \url{https://arxiv.org/abs/2601.07055}.

\bibitem[Zhang et~al.(2025)Zhang, Huang, Song, Zhu, Zhang, Zhao, and Zhao]{criticsearch}
Yaocheng Zhang, Haohuan Huang, Zijun Song, Yuanheng Zhu, Qichao Zhang, Zijie Zhao, and Dongbin Zhao.
\newblock {CriticSearch: Fine-Grained Credit Assignment for Search Agents via a Retrospective Critic}, 2025.
\newblock URL \url{https://arxiv.org/abs/2511.12159}.

\bibitem[Zheng et~al.(2025{\natexlab{a}})Zheng, An, Wang, Wang, and Wu]{zheng-etal-2025-stepsearch}
Xuhui Zheng, Kang An, Ziliang Wang, Yuhang Wang, and Yichao Wu.
\newblock {S}tep{S}earch: Igniting {LLM}s search ability via step-wise proximal policy optimization.
\newblock In Christos Christodoulopoulos, Tanmoy Chakraborty, Carolyn Rose, and Violet Peng (eds.), \emph{Proceedings of the 2025 Conference on Empirical Methods in Natural Language Processing}, pp.\  21805--21830, Suzhou, China, November 2025{\natexlab{a}}. Association for Computational Linguistics.
\newblock ISBN 979-8-89176-332-6.
\newblock \doi{10.18653/v1/2025.emnlp-main.1106}.
\newblock URL \url{https://aclanthology.org/2025.emnlp-main.1106/}.

\bibitem[Zheng et~al.(2025{\natexlab{b}})Zheng, Fu, Hu, Cai, Ye, Lu, and Liu]{deepresearcher}
Yuxiang Zheng, Dayuan Fu, Xiangkun Hu, Xiaojie Cai, Lyumanshan Ye, Pengrui Lu, and Pengfei Liu.
\newblock {D}eep{R}esearcher: Scaling deep research via reinforcement learning in real-world environments.
\newblock In Christos Christodoulopoulos, Tanmoy Chakraborty, Carolyn Rose, and Violet Peng (eds.), \emph{Proceedings of the 2025 Conference on Empirical Methods in Natural Language Processing}, pp.\  414--431, Suzhou, China, November 2025{\natexlab{b}}. Association for Computational Linguistics.
\newblock ISBN 979-8-89176-332-6.
\newblock \doi{10.18653/v1/2025.emnlp-main.22}.
\newblock URL \url{https://aclanthology.org/2025.emnlp-main.22/}.

\end{thebibliography}
\bibliographystyle{colm2026}

\newpage
\appendix
\section{Extended Related Work}\label{sec:extended-related-work}

This section contains a detailed overview of related search agent papers:

\paragraph{Search-o1 \citep{search_o1}} One of the first works to incorporate multi-turn retrieval and reasoning by converting the retrieved documents into focused reasoning steps that integrate the external knowledge, while maintaining the logical flow of the reasoning chain. The authors used the QwQ-32B-Preview as the base model, employed the Bing Web Search for web search and Jina Reader API to fetch the content of the web page. The QwQ-32B-Preview was not fine-tuned using reinforcement learning, instead was naively used for multi-turn retrieval and reasoning on open-domain question-answering tasks on Wikipedia.

\paragraph{Search-R1 \citep{search_r1}} One of the earliest works of training LLMs with multi-turn search engine calls and LLM response rollout using reinforcement learning with PPO. 
Search-R1 used two LLMs as the base model: Qwen2.5-3B/7B (Base/Instruct)~\citep{qwen25} and trained them on two Wikipedia datasets: NQ~\citep{kwiatkowski-etal-2019-natural} and HotpotQA~\citep{yang-etal-2018-hotpotqa}. 
Both datasets contain queries that can be answered using 1--2 hops of information searches, the models are trained using a single RL stage containing a maximum of 2 turns and rollout 5 responses per prompt (GRPO). For evaluation, Search-R1 retrieved top 3 documents from the 2018 December Wikipedia corpus containing chunked documents with a maximum of 512 tokens~\citep{karpukhin-etal-2020-dense} with \texttt{e5-base-v2}~\citep{wang:2022}\footnote{\href{https://huggingface.co/intfloat/e5-base-v2}{\texttt{https://huggingface.co/intfloat/e5-base-v2}}} as the dense retrieval model with 110M parameters, 12 layers and an embedding size of 768 dimensions. They observed PPO to be stable than GRPO for training models in their experiments.

\paragraph{ReSearch \citep{reSearch}} ReSearch is similar to Search-R1. ReSearch trains Qwen2.5-7B/32B (Base/Instruct) LLMs using reinforcement learning with GRPO. ReSearch modified the Search-R1 reward by adding a format-based reward to teach the model to follow the required format. Models were trained on the training dataset of MuSiQue~\citep{musique} with 5 rollouts. The evaluation setting is identical to Search-R1, with the only caveat of retrieving top-5 documents for each query.

\paragraph{R1-Searcher \citep{r1_searcher}} R1-Searcher is similar to both ReSearch and Search-R1. R1-Searcher trained Qwen2.5-7B (Base) and Llama-3.1-8B-Instruct LLMs using reinforcement learning with GRPO. They include a retrieval reward in the first-stage of RL training, followed by the answer reward in the second stage. They train model on two datasets: Hotpotqa~\citep{yang-etal-2018-hotpotqa} and 2WikiMultiHopQA~\citep{2wikimultihopqa} with 16 rollouts. The evaluation setup is different from previous setups, they incorporate the English Wikipedia corpus provided by KILT~\citep{petroni-etal-2021-kilt} and use the \texttt{BGE-large-en-v1.5}\footnote{\href{https://huggingface.co/BAAI/bge-large-en-v1.5}{\texttt{https://huggingface.co/BAAI/bge-large-en-v1.5}}} as the dense retriever model, retrieving top-8 documents (except Bamboogle, where they employ Google web search API).

\paragraph{DeepResearcher \citep{deepresearcher}} One of the first works of training LLMs with multi-turn search engine calls from a live search engine, such as Google web search API. They fine-tuned the Qwen2.5-7B Instruct model and trained on four datasets: NQ, TriviaQA, HotpotQA and 2WikiMultiHopQA with a distribution ratio of (1:~1:~3:~3) with a focus on multi-hop scenarios, using GRPO with 16 rollouts and a maximum turn size of 10. The evaluation was done with web search on a subset of 512 evaluation samples; making the reproducibility difficult, as the subsets are not publicly available.

\paragraph{SimpleDeepSearcher \citep{simpledeepsearcher}} One of the first to show that SFT with reasoning trajectories works better than RL for training models. They SFT train four models: Qwen2.57B-Instruct, Qwen2.5-32B-Instruct, DeepseekDistilled-Qwen2.5-32B, and QwQ-32B with live web-search. Similar to DeepResearcher, they evaluate the models on 500 randomly sampled evaluation samples.

\paragraph{InForage \citep{inforage}} InForage introduces three reward mechanisms to incentivize comprehensive reasoning, by including an information gain reward, rewarding retrieval steps with valuable evidence and an efficiency penalty discouraging prolonged reasoning. InForage constructed their own dataset, where human annotators begin with a seed factoid claim and used Google Search API to extend the claim by looking at selected web-pages and expand the context around it. They manually annotated for 500 queries. Next, they expanded it automatically with GPT-4o for around 20K training samples. The Qwen2.5-Instruct 3B and 7B were used as the base models for training. Each model first went through a supervised fine-tuning stage (SFT) and further a reinforcement learning stage with PPO trained on a mixture of the self-constructed training dataset, NQ and HotpotQA. The maximum number of turns during training was 6, with cached Google search results and scraped full content of webpages. The inference setting is similar to Search-R1 except they use the \texttt{BGE-M3} model~\citep{bge-m3}\footnote{\href{https://huggingface.co/BAAI/bge-m3}{\texttt{https://huggingface.co/BAAI/bge-m3}}} as the dense retrieval model.

\paragraph{InfoSeek \citep{infoseek}} InfoSeek focuses on automated data synthesis, by constructing a training dataset to perform on reasoning-intensive queries using hierarchical constraints requiring raw entities and their relations. They fine-tune models using SFT via rejection sampling for the first stage of training followed by RL training in second stage. The train the Qwen2.5-3B-Instruct with 5 rollouts, and a maximum turn size of 10, with the search-engine retrieving top-5 contents. They fine-tune with GRPO on 17K harder samples (15K InfoSeek, 2K NQ and HotpotQA) in the second stage. The evaluation setting uses the Wikipedia corpus from 2025 segmented into chunks of 512 tokens and retrieve using the \texttt{BGE-M3} retrieval model with top-5 documents retrieved for each query.

\paragraph{WebSailor \citep{websailor}} WebSailor also focuses on automated data synthesis, by synthesizing high-uncertainty web navigation tasks to train specialized agents. WebSailor's approach is focused on building knowledge graphs and subgraph sampling and obfuscation. They train four models: Qwen2.5-3B/7B/32B/72B (Instruct) using SFT with rejection sampling and further train in the next stage using Duplicating Sampling Policy Optimization (DUPO) with 8 rollouts. No other training details were provided. They utilize two tools:\ 
(1) \emph{search} for returning top-10 search results with Google Search, and (2) \emph{visit} to summarize the information for each webpage using Qwen2.5-72B as the summary model.

\paragraph{WebShaper \citep{webshaper}} WebShaper introduces a formalization-guided framework for data synthesis, by linking relationships between entities in order to expand existing questions. After constructing information seeking pairs, they use an expander agent to enhance the original question via iterative refinement. They train on three Qwen model variants: Qwen2.5-32B/72B and QwQ-32B using SFT with rejection sampling. Next, they train their models with DAPO with 8 rollouts. The temperature is set as 1.0, batch size as 128, and the learning rate as 1e-6.

\section{Search Agent Prompt}\label{sec:search-agent-prompt}

\begin{figure*}[t]
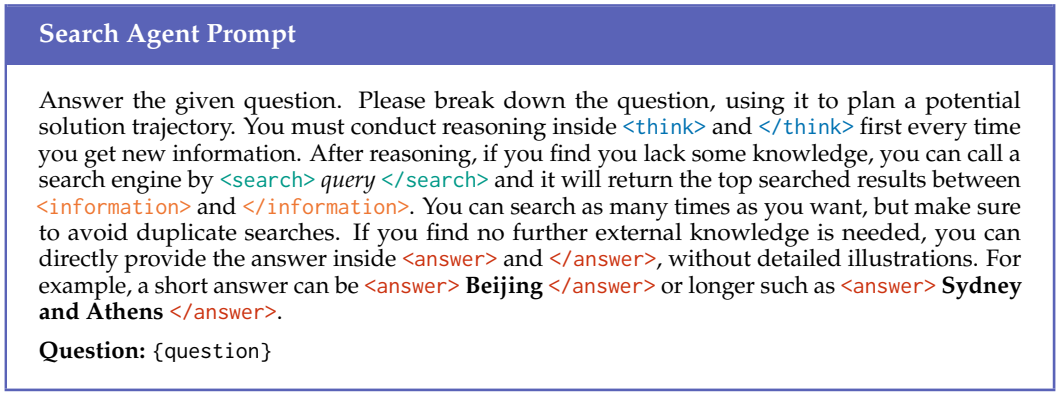

    \centering
    \small
    \vspace{-0.5em}
    \begin{mdframed}[
        font=\small,
        roundcorner=6pt,
        linecolor=odysseyblue,
        linewidth=1pt,
        innerleftmargin=12pt,
        innerrightmargin=12pt,
        innertopmargin=10pt,
        innerbottommargin=10pt,
        frametitle={\color{white}\bfseries\normalsize Search Agent Prompt},
        frametitlebackgroundcolor=odysseyblue,
        frametitleaboveskip=6pt,
        frametitlebelowskip=6pt
    ]
    Answer the given question. Please break down the question, using it to plan a potential solution trajectory.
    You must conduct reasoning inside
    \textcolor{orb_thinkcolor}{\texttt{<think>}} and \textcolor{orb_thinkcolor}{\texttt{</think>}}
    first every time you get new information.
    After reasoning, if you find you lack some knowledge, you can call a search engine by
    \textcolor{orb_searchcolor}{\texttt{<search>}} \textit{query} \textcolor{orb_searchcolor}{\texttt{</search>}}
    and it will return the top searched results between
    \textcolor{orb_infocolor}{\texttt{<information>}} and \textcolor{orb_infocolor}{\texttt{</information>}}.
    You can search as many times as you want, but make sure to avoid duplicate searches.
    If you find no further external knowledge is needed, you can directly provide the answer inside
    \textcolor{orb_anscolor}{\texttt{<answer>}} and \textcolor{orb_anscolor}{\texttt{</answer>}},
    without detailed illustrations.
    For example, a short answer can be
    \textcolor{orb_anscolor}{\texttt{<answer>}} \textbf{Beijing} \textcolor{orb_anscolor}{\texttt{</answer>}}
    or longer such as
    \textcolor{orb_anscolor}{\texttt{<answer>}} \textbf{Sydney and Athens} \textcolor{orb_anscolor}{\texttt{</answer>}}.
    \\[0.5em]
    \textbf{Question:} \texttt{\{question\}}
    \end{mdframed}
    \vspace{-0.5em}
    \caption{Prompt template used for search agent training in \orbit. The agent interleaves
    \orbtagthink{reasoning} blocks with \orbtagsearch{query} calls; retrieved passages are returned as \orbtaginfo{documents}, and the trajectory terminates when the model emits a final \orbtagans{answer}.}
    \label{fig:orbit-prompt-template}
\end{figure*}

\autoref{fig:orbit-prompt-template} shows the prompt template used to train \orbit-4B via GRPO in our work.
The template instructs \orbit-4B to first \emph{plan} a solution trajectory by decomposing the question before issuing any search calls, a design choice motivated by the multi-hop, reasoning-intensive nature of \orbit queries.

Concretely, the agent alternates between internal deliberation enclosed in \orbtagthink{} tags and targeted web queries wrapped in \orbtagsearch{} tags.
Each query is routed to DDGS web retrieval server (top-$k$=5 documents) whose results are injected back (containing the title and snippet information) into the context as \orbtaginfo{} blocks.
The agent may issue arbitrarily many search calls per trajectory but is explicitly discouraged from repeating identical queries, which would otherwise dominate the turn budget without contributing new evidence.
The trajectory concludes when the model emits a \orbtagans{} tag, whose content is matched against the ground-truth answer using exact-match (EM) scoring to produce the GRPO reward signal.

\section{\orbit-4B Training Hyperparameters}\label{sec:orbit-hyperparams}

We train the \orbit-4B search agent using the \texttt{verl-tool} framework~\citep{2025verltool}, which extends VeRL with multi-turn, tool-augmented RL training capabilities.
Training uses GRPO~\citep{shao:2024} with a live DDGS-based web retriever (top-$k$=5, parallel fan-out across \texttt{google}, \texttt{brave}, \texttt{bing}, \texttt{wikipedia}, and \texttt{grokipedia} backends).
Rollouts are executed asynchronously via vLLM (v1 engine), with FSDP parameter and optimizer offloading to support the 8{,}192 token context window on a single 4$\times$H100 SXM5 node.
Table~\ref{tab:orbit-hyperparams} summarizes the full configuration.

\begin{table}[htbp]
\centering
\small
\resizebox{0.6\textwidth}{!}{%
\begin{tabular}{@{}llc@{}}
\toprule
\textbf{Module} & \textbf{Hyper-parameter} & \textbf{Value} \\
\midrule
\addlinespace[0.4ex]
\textbf{Data}
  & Max observation (search result) length & 1{,}024 \\
  & Max response length                    & 8{,}192 \\
  & Max prompt length                      & 2{,}048 \\
  & Max action length                      & 2{,}048 \\
  & Retriever top-$k$                      & 5 (using DDGS) \\
\midrule
\textbf{Actor}
  & Training batch size                    & 256 \\
  & Mini-batch size                        & 32 \\
  & Learning rate                          & $1\times10^{-6}$ \\
  & LR warmup steps                        & 10 \\
  & KL coefficient ($\beta$)               & 0.0 \\
  & KL loss type                           & \texttt{low\_var\_kl} \\
  & Entropy coefficient                    & 0.0 \\
  & Parallelism strategy                   & FSDP \\
\midrule
\textbf{Rollout}
  & Max turns or search actions            & 5 \\
  & Group size $G$ (rollouts per sample)   & 8 \\
  & Temperature                            & 1.0 \\
  & Top-$p$                                & 1.0 \\
  & vLLM GPU memory utilization            & 0.6 \\
  & vLLM max model length                  & 8{,}192 \\
  & Rollout mode                           & Async (vLLM v1) \\
\midrule
\textbf{Reward}
  & Reward function                        & Exact Match (EM) \\
  & Mask observations in loss              & Yes \\
\bottomrule
\end{tabular}
}
\caption{Hyperparameters used to train \orbit-4B and other search agents (Search-R1-4B and InfoSeeker-4B) with GRPO via the \texttt{verl-tool} framework~\citep{2025verltool}.}
\label{tab:orbit-hyperparams}
\end{table}

\section{Extended Comparison using the Search-R1 Evaluation Setup}
\label{app:extended-comparison}

We extend the \orbit-4B results with a broader set of search agent baselines, by evaluating search agents trained on Qwen2.5-3B and our own baselines on Qwen3-4B by reproducing the Search-R1 evaluation setup~\citep{search_r1}.
We retrieve passages from the 2018 Wikipedia dump~\citep{karpukhin-etal-2020-dense} using the E5-base-v2 retriever~\citep{wang:2022}, retrieving top-3 passages at each turn.
Results for search agents trained using Qwen2.5-3B as the base model are taken directly from SAPO \citep{li2026improvingsearchagentline}, that evaluated these search agents in the Search-R1 setting~\citep{search_r1}.

\paragraph{Extended Results.} \autoref{tab:extended-comparison} provides the EM accuracy of search agents evaluated by reproducing the Search-R1 evaluation setup~\citep{search_r1}, i.e., using the 2018 Wikipedia dump, E5-base-v2 retriever providing top-3 documents at each turn. 
From the results, we observe that \orbit-4B achieves 44.9 EM accuracy outperforming InfoSeeker-4B by 3.6 points and Search-R1-4B by 6.8 points, on average. In addition, \orbit-4B outperforms existing search agents trained with Qwen2.5-3B (Base or Instruct), demonstrating that using the recent and newer Qwen3-4B as the base model can outperform search agents trained with extensive training regimes with Qwen2.5-3B.

\begin{table*}[t]
\centering
\small
\setlength{\tabcolsep}{4.5pt}
\renewcommand{\arraystretch}{1.05}
\resizebox{\linewidth}{!}{
\begin{tabular}{lcccccccc}
\toprule
\multirow{2}{*}{\textbf{Methods}} & \multicolumn{3}{c}{\textbf{Single-Hop QA}} & \multicolumn{4}{c}{\textbf{Multi-Hop QA}} & \multirow{2}{*}{\textbf{Avg.~7}} \\
\cmidrule(lr){2-4} \cmidrule(lr){5-8}
 & NQ & TQA & PopQA & HQA & 2Wiki & MSQ & Bamb & \\
\midrule

\rowcolor{gray!10} \multicolumn{9}{l}{\textbf{Search Agents} (Base: Qwen2.5-3B)} \\
 Search-o1 \citep{search_o1}                   & 23.8 & 47.2 & 26.2 & 22.1 & 21.8 &  5.4 & 32.0 & 25.5 \\
 Search-R1-Instruct \citep{search_r1}          & 39.7 & 56.5 & 39.1 & 33.1 & 31.0 & 12.4 & 23.2 & 33.6 \\
 Search-R1-Base \citep{search_r1}              & 42.1 & 58.3 & 41.3 & 29.7 & 27.4 &  6.6 & 12.8 & 31.2 \\
 ReSearch-Instruct \citep{reSearch}            & 36.5 & 57.1 & 39.5 & 35.1 & 27.2 &  9.5 & 26.6 & 33.1 \\
 ReSearch-Base \citep{reSearch}                & 42.7 & 59.7 & 43.0 & 30.5 & 27.2 &  7.4 & 12.8 & 31.9 \\
 Dr.~Zero-3B \citep{dr-zero:2026} & 39.7 & 57.2 & 43.1 & 29.8 & 29.1 & 9.1 & 20.0 & 32.6 \\
 ZeroSearch-Base \citep{zerosearchbase:2025}     & 43.0 & 61.6 & 41.4 & 33.8 & 34.6 & 13.0 & 13.9 & 34.5 \\
 StepSearch-Base \citep{zheng-etal-2025-stepsearch}    & \nmark & \nmark & \nmark & 32.9 & 33.9 & 18.1 & 32.8 & \nmark \\
 EXSEARCH-Base \citep{shi2025iterative}         & 36.8 & \nmark & \nmark & 42.2 & 37.2 & 13.8 & \nmark & \nmark \\
 $O^2$-Searcher \citep{mei:2025}     & 44.4 & 59.7 & 42.9 & 38.8 & 37.4 & 16.0 & 34.4 & 39.1 \\
 AutoRefine-Instruct \citep{shi2025search} & 43.6 & 59.7 & 44.7 & 40.4 & 38.0 & 16.9 & 33.6 & 39.6 \\
 AutoRefine-Base \citep{shi2025search}     & 46.7 & 62.0 & 45.0 & 40.5 & 39.3 & 15.7 & 34.4 & 40.5 \\
 InForage \citep{inforage}                     & 42.1 & 59.7 & 45.2 & 40.9 & 42.8 & 17.2 & 36.0 & 40.5 \\
 CriticSearch \citep{criticsearch}             & \nmark & \nmark & \nmark & 41.4 & 40.9 & 18.0 & 36.8 & \nmark \\
 SE-Search \citep{li2026sesearchselfevolvingsearchagent}              & \textcolor{paperblue}{\textbf{47.5}} & 62.4 & 42.3 & 45.0 & 36.1 & 18.3 & 42.4 & 42.0 \\
 
 SAPO-3B-Base \citep{li2026improvingsearchagentline}                 & 47.4 & 63.0 & 44.9 & 44.9 & 41.2 & 19.6 & 42.4 & 43.3 \\
 SAPO-3B-Instruct \citep{li2026improvingsearchagentline}             & 46.9 & 62.9 & 45.7 & \textcolor{paperblue}{\textbf{45.6}} & 44.8 & \textcolor{paperblue}{\textbf{20.3}} & 43.2 & 44.2 \\
\midrule

\rowcolor{odysseyrow} \multicolumn{9}{l}{\textcolor{paperblue}{\textbf{Search Agents} (Base: Qwen3-4B)}} \\
Search-R1-4B$^\dagger$ & 37.9 & 58.8 & 40.7 & 34.3 & 37.1 & 13.0 & 44.8 & 38.1 \\

InfoSeeker-4B$^\dagger$ & 38.1 & 63.8 & 42.1 & 40.4 & 41.1 & 15.7 & 48.0 & 41.3 \\
\cdashline{1-9}
\rowcolor{odysseyrow}
\textcolor{paperblue}{\textbf{\orbit-4B$^\dagger$}} & 44.7 & \textcolor{paperblue}{\textbf{66.3}} & \textcolor{paperblue}{\textbf{47.1}} & 42.3 & \textcolor{paperblue}{\textbf{45.9}} & 19.4 & \textcolor{paperblue}{\textbf{48.8}} & \textcolor{paperblue}{\textbf{44.9}} \\
\bottomrule
\end{tabular}
}
\caption{Accuracy comparison of search agents $\leq$ 4B parameters on single-hop and multi-hop QA benchmarks using the \textbf{Search-R1} evaluation setup~\citep{search_r1}, with the \textbf{2018 Wikipedia dump}~\citep{karpukhin-etal-2020-dense}, \textbf{E5-base-v2} retriever~\citep{wang:2022} serving as the retrieval engine retrieving \textbf{top-3} documents for every search query. Qwen2.5-3B results are taken from SAPO~\citep{li2026improvingsearchagentline}.
($\dagger$) denotes search agents trained in our work using Qwen3-4B as the base model. Best scores are highlighted in \textcolor{paperblue}{\textbf{bold}}.}
\label{tab:extended-comparison}
\end{table*}

\section{\orbit Dataset Examples}
\label{app:domain-examples}

We present one representative training example from each of the 15 knowledge domains in \orbit (dataset example for the domain \emph{TV Shows \& Movies} is already shown in \autoref{tab:odyssey-example}).
Colored spans highlight the individual factual clues embedded in each question; the verification
section confirms every clue with the key supporting fact (shown in \textcolor{blue}{blue}).

\clearpage

\begin{table}[t]
\centering
\footnotesize
\setlength{\tabcolsep}{6pt}
\renewcommand{\arraystretch}{1}
\begin{tabularx}{\columnwidth}{@{} p{0.24\columnwidth} X @{}}
\toprule
\textbf{Answer Type} & \textbf{Definition} \\
\midrule
\textbf{Temporal} & Calendar date, year, month, decade, or explicitly time-anchored period. \\ \midrule
\textbf{Numeric / Quantitative} & Numerical value, measurement, ratio, duration, or rate, with or without units. \\ \midrule
\textbf{Geographic Location} & Physical place, region, landmark, or spatially identifiable location on Earth. \\ \midrule
\textbf{Person} & Named individual (historical, contemporary, or fictional). \\ \midrule
\textbf{Organization / Institution} & Named group, body, office, society, standardization entity, or collective actor. \\ \midrule
\textbf{Event / Historical Phenomenon} & Named occurrence, movement, crisis, or episode that happens in time. \\ \midrule
\textbf{Scientific / Technical Concept} & Formally defined concept, theory, method, model, or phenomenon. \\ \midrule
\textbf{Named Artifact / System} & Specific non-human entity such as a tool, system, standard, protocol, document, product, currency, or legal instrument. \\ \midrule
\textbf{Property / Relationship} & Characteristic, state, quality, constraint, or relational descriptor of an entity. \\
\bottomrule
\end{tabularx}
\caption{Definitions of answer types used for categorizing answers in the \orbit dataset.}
\label{tab:answer-types}
\end{table}
\begin{figure}[t]
\centering
\begin{mdframed}[
    font=\scriptsize,
    roundcorner=10pt,
    linecolor=odysseyblue,
    linewidth=1pt,
    innerleftmargin=10pt,
    innerrightmargin=10pt,
    innertopmargin=10pt,
    innerbottommargin=10pt,
    frametitle={\color{white}\bfseries\normalsize Question--Answer Pair Generation Prompt},
    frametitlebackgroundcolor=odysseyblue,
    frametitleaboveskip=6pt,
    frametitlebelowskip=6pt
]

\textbf{SYSTEM:} You are an expert multi-hop factoid question creator. You should create complex or inverted questions containing answers that are easy to verify, but hard to solve for a given seed, along with the answer, a verification checklist, and a list of evidence URLs sufficient to justify the answer.\\[0.5em]

\textbf{TASK:} The question-answer pair generated should satisfy the conditions below:
\begin{itemize}[leftmargin=*, itemsep=0pt]
    \item The question should have a unique answer.
    \item The answer should be a short fact (a term or a short phrase).
    \item The question should be very difficult to answer without the list of correct evidence URLs, requiring multiple web searches.
    \item The model answer should be easily verifiable using the evidence URLs and should appear as a substring in the final evidence.
    \item The question should be phrased so that each evidence does not give away too much information about the model answer so that the student can determine the model answer without finding all the evidences you've listed. That is, each evidence should be absolutely necessary to answer the question correctly.
    \item The question should be deep enough that it requires at least 5 evidence to answer the question correctly.
    \item You should not mention any proper nouns in the question as it gives away too much information about the model answer. Instead, describe the proper noun's properties in a way that is still enough to uniquely identify the proper noun. \\[0.5em]
\end{itemize} 

\textbf{PROCEDURE:} You would typically start with a \emph{seed} (could be a person, event, or artifact) that the user will provide. Find several characteristics within a large search space using the \emph{seed}, and create a question from them.\\[0.5em]

\textbf{REQUIRED OUTPUT:} Please make sure you look into the seed carefully and think first before you provide your output in XML format. Make sure the \emph{seed} is actually hidden in the question to make it challenging and your question should not be too verbose or easy to answer. Follow the format below by providing a verified answer to the inverted question, a verification checklist containing citations containing evidence for each verification, and a list of evidence URLs:

\begin{verbatim}
<output>
    <inverted_question>your inverted query</inverted_question>
    <answer>the answer to the inverted query</answer>
    <verification_checklist>
        <item>bullet point 1 :cite[1]</item>
        <item>bullet point 2 :cite[1]</item>
        <item>bullet point 3 :cite[7]</item>
        etc.
    </verification_checklist>
    <evidence_urls>[1]: https://xxxx..., [2]: https://yyy..., [7]: https://zzzz...</evidence_urls>
</output> 
\end{verbatim} 
\vspace{2mm}

\textbf{EXAMPLE:}  \{\texttt{Add an example here}\} \\

\textbf{NOTE:} Make sure the \emph{seed} is actually hidden in the question to make it challenging and also do not use the \emph{seed} as your answer! Further, to ensure that your question is difficult enough, you should try to answer the question yourself from scratch, and see if you can answer it confidently by using less than 5 searches. If so, revise your question and try again. \\

\textbf{SEED:} \texttt{\{seed\}}
\end{mdframed}
\vspace{-0.2cm}
\caption{Prompt template for question--answer pair generation for a given input seed as inspiration and shuffled exemplars.}
\label{fig:prompt-multihop}
\vspace{-0.2cm}
\end{figure}
\begin{figure}[t!]
\centering
\begin{mdframed}[
    font=\scriptsize,
    roundcorner=10pt,
    linecolor=odysseyblue,
    linewidth=1pt,
    innerleftmargin=10pt,
    innerrightmargin=10pt,
    innertopmargin=10pt,
    innerbottommargin=10pt,
    frametitle={\color{white}\bfseries\normalsize Self-Verification Prompt},
    frametitlebackgroundcolor=odysseyblue,
    frametitleaboveskip=6pt,
    frametitlebelowskip=6pt
]

\textbf{QUESTION:} \{\texttt{question}\} \\
\textbf{ANSWER:} \{\texttt{answer}\} \\

Given the question and answer, conduct a full web and Wikipedia search to retrieve the pages necessary to answer the question. Next, come up with a verification list of criteria, citing each URL for each criterion. 
Finally, provide a revised answer to the question (a short string) if needed, and lastly, list all the sources you cite with each URL and which all verification statement it supports! \\[0.5em]

Please do the web search first!

\end{mdframed}
\vspace{-0.2cm}
\caption{Prompt template for self-verification given the input question and answer.}
\label{fig:prompt-self-verification}
\vspace{-0.2cm}
\end{figure}
\begin{figure}[t!]
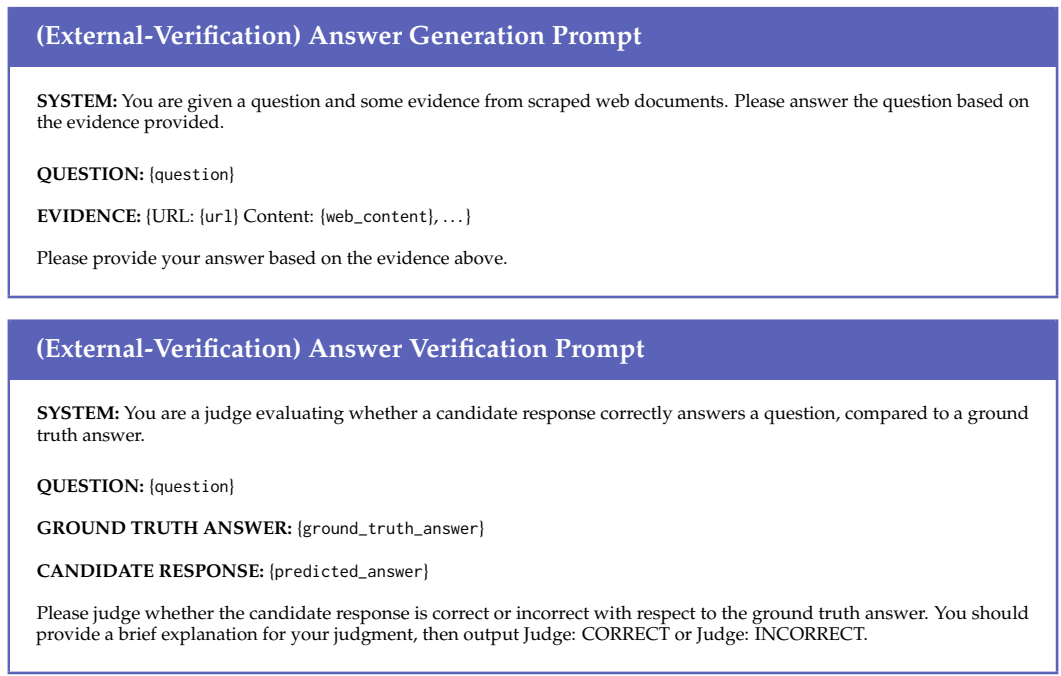

\centering
\begin{mdframed}[
    font=\scriptsize,
    roundcorner=10pt,
    linecolor=odysseyblue,
    linewidth=1pt,
    innerleftmargin=10pt,
    innerrightmargin=10pt,
    innertopmargin=10pt,
    innerbottommargin=10pt,
    frametitle={\color{white}\bfseries\normalsize (External-Verification) Answer Generation Prompt},
    frametitlebackgroundcolor=odysseyblue,
    frametitleaboveskip=6pt,
    frametitlebelowskip=6pt
]

\textbf{SYSTEM:} You are given a question and some evidence from scraped web documents. Please answer the question based on the evidence provided. \\[0.5em]

\textbf{QUESTION:} \{\texttt{question}\} \\

\textbf{EVIDENCE:} \{URL: \{\texttt{url}\} Content: \{\texttt{web\_content}\}, \ldots \} \\

Please provide your answer based on the evidence above.
\end{mdframed}
\begin{mdframed}[
    font=\scriptsize,
    roundcorner=10pt,
    linecolor=odysseyblue,
    linewidth=1pt,
    innerleftmargin=10pt,
    innerrightmargin=10pt,
    innertopmargin=10pt,
    innerbottommargin=10pt,
    frametitle={\color{white}\bfseries\normalsize (External-Verification) Answer Verification Prompt},
    frametitlebackgroundcolor=odysseyblue,
    frametitleaboveskip=6pt,
    frametitlebelowskip=6pt
]

\textbf{SYSTEM:} You are a judge evaluating whether a candidate response correctly answers a question, compared to a ground truth answer. \\[0.5em]

\textbf{QUESTION:} \{\texttt{question}\} \\

\textbf{GROUND TRUTH ANSWER:} \{\texttt{ground\_truth\_answer}\} \\

\textbf{CANDIDATE RESPONSE:} \{\texttt{predicted\_answer}\} \\

Please judge whether the candidate response is correct or incorrect with respect to the ground truth answer. You should provide a brief explanation for your judgment, then output Judge: CORRECT or Judge: INCORRECT.
\end{mdframed}
\vspace{-0.2cm}
\caption{Prompt template for external verification in \orbit, first the LLM independently generates an answer for the question given the evidence context. Next, the same LLM evaluates whether the predicted answer matches with the ground truth answer, and outputs either correct or incorrect.}
\label{fig:prompt-external-generation}
\vspace{-0.2cm}
\end{figure}
\begin{table*}[t!]
    \centering
    \small
    \vspace{-0.5em}

    \begin{tabularx}{\linewidth}{X}
    \toprule
    \rowcolor{odysseyrow} \multicolumn{1}{c}{\textbf{Case Study on FRAMES}~\citep{krishna-etal-2025-fact}}
    \\ \midrule
    \textbf{Question:} What compass direction (of the 4 cardinal and 4 ordinal directions) is the capital city of the state which houses the Jackson Hole Mountain Resort in relation to the centre of the state? \\
    \textbf{Ground Truth:} [``Southeast''] \\
    \midrule
    \textbf{Model Output: \orbit-4B} \\
    \midrule

    \orbtagthink{%
        The question requires three reasoning hops: (1)~identify the state that contains Jackson Hole Mountain Resort, (2)~find the capital of that state, and (3)~determine the capital's compass direction relative to the state's geographic centre.
        I recall that Jackson Hole lies in Wyoming; the capital is likely Cheyenne, but I should confirm.%
    }
    \vspace{0.4em} \\

    \orbtagsearch{capital of Wyoming}
    \vspace{0.2em} \\

    \orbtaginfo{%
        \textbf{Doc 1 (Title: ``Wyoming State Capitol''):} The Wyoming State Capitol is located in the city of \textbf{Cheyenne}.
        Cheyenne is the \textbf{seat of government for the U.S.\ state of Wyoming}\ldots\\
        \textbf{Doc 2 (Title: ``Cheyenne, Wyoming''):} \textbf{Cheyenne} \ldots is the \textbf{capital and most populous city of the U.S.\ state of Wyoming}, as well as the county seat of Laramie County, with 65{,}132 residents per the 2020 census\ldots\\
        \textbf{Doc 3 (Title: ``Wyoming''):} The \textbf{state capital and most populous city is Cheyenne}\ldots%
    } \\
    \midrule

    \orbtagthink{%
        Confirmed: the state is Wyoming and its capital is Cheyenne.
        I now need the geographic centre of Wyoming so I can determine the direction from that centre to Cheyenne.%
    }
    \vspace{0.4em} \\

    \orbtagsearch{geographic center of Wyoming}
    \vspace{0.2em} \\

    \orbtaginfo{%
        \textbf{Doc 3 (Title: ``Outline of Wyoming''):} Geography of Wyoming --- Wyoming is a U.S.\ state \ldots located in the Western United States \ldots\\
        \textbf{Doc 4 (Title: ``Geography of Wyoming''):} The Wind River Range in the \textbf{west central part} of the state\ldots The Big Horn Mountains in the \textbf{north central portion}\ldots%
    } \\
    \midrule

    \orbtagthink{%
        These results describe Wyoming's geography but do not give the geographic centre explicitly.
        I already know Wyoming is a roughly rectangular state; the centre is therefore approximately mid-state.
        Let me ask directly where Cheyenne sits relative to the rest of the state---that should encode the direction implicitly.%
    }
    \vspace{0.4em} \\

    \orbtagsearch{direction of Cheyenne, Wyoming relative to center of state}
    \vspace{0.2em} \\

    \orbtaginfo{%
        \textbf{Doc 1 (Title: ``Cheyenne, Wyoming''):} \textbf{Lying near the southeast corner of the state}, Cheyenne is one of the least centrally located state capitals in the nation (together with cities such as Carson City, Nevada; Juneau, Alaska; Tallahassee, Florida; Topeka, Kansas; and Trenton, New Jersey).\\
        \textbf{Doc 3 (Title: ``Laramie County, Wyoming''):} The county seat is Cheyenne, the state capital. The county \textbf{lies west of the Nebraska state line and north of the Colorado state line} [i.e.\ in the southeast corner of Wyoming].%
    } \\
    \midrule

    \orbtagthink{%
        The retrieved document explicitly states that Cheyenne lies near the \emph{southeast corner} of Wyoming.
        The state's geographic centre therefore lies to the northwest of Cheyenne, which equivalently means Cheyenne is to the \textbf{southeast} of the centre.
        The answer is \textbf{southeast}.%
    }
    \vspace{0.4em} \\

    \orbtagans{Southeast} \\

    \bottomrule
    \end{tabularx}
    \caption{Case study of \orbit-4B evaluated on FRAMES~\citep{krishna-etal-2025-fact}. The search agent must chain geographic facts across multiple hops; key evidence extracted from retrieved documents is \textbf{bolded}.}
    \label{tab:orbit-case-study}
\end{table*}

\newpage
\begin{table*}[t!]
    \centering
    \small
    \vspace{-0.5em}

    \begin{tabularx}{\linewidth}{X}
    \toprule
    \rowcolor{odysseyrow} \multicolumn{1}{c}{\textbf{Case Study on \orbit during GRPO Training}}
    \\ \midrule
    \textbf{Question:} In which 1957 nature essay collection does a chapter describing a boy's unexpected discovery of a sunfish in a city creek---illustrating how preconceptions blind observers to novel phenomena---appear, written by an anthropologist-naturalist who later became president of the American Institute of Human Paleontology? \\
    \textbf{Ground Truth:} [``The Immense Journey''] \\
    \midrule
    \textbf{Model Output: \orbit-4B} \\
    \midrule

    \orbtagthink{%
        I need a 1957 nature essay collection with a sunfish chapter, written by an anthropologist-naturalist who later led the American Institute of Human Paleontology.
        This is a multi-hop question: (1)~identify the author via the institute clue, (2)~find the 1957 collection they wrote, and (3)~confirm the sunfish chapter.
        Let me first search by the essay's contents.%
    }
    \vspace{0.4em} \\

    \orbtagsearch{1957 nature essay collection sunfish city creek}
    \vspace{0.2em} \\

    \orbtaginfo{%
        \textbf{Doc 1 (Title: ``1957 --- Wikipedia''):} 1957 (MCMLVII) was a common year starting on Tuesday of the Gregorian calendar\ldots\\
        \textbf{Doc 2 (Title: ``Historical Events in 1957''):} Jan 22, 2013 --- Historical events from year 1957. Learn about 514 famous, scandalous and important events\ldots\\
        \textbf{Doc 5 (Title: ``1957 Events \& Facts''):} MAJOR EVENTS: Soviet Union inaugurates the ``Space Age'' by launching Sputnik~I\ldots%
    } \\
    \midrule

    \orbtagthink{%
        Those results are not relevant---only generic 1957 almanac pages. Searching by essay content alone seems too obscure for the retriever.
        Let me try a more direct query about the specific episode.%
    }
    \vspace{0.4em} \\

    \orbtagsearch{sunfish in a city creek essay 1957}
    \vspace{0.2em} \\

    \orbtaginfo{%
        \textbf{Doc 1 (Title: ``Sunfish Age and Hull Identification --- SailingForums.com''):} \ldots archives the full history of the Sunfish \emph{sailboat} builders over the years\ldots\\
        \textbf{Doc 3 (Title: ``Sunfish vs Minifish? --- SailingForums.com''):} \ldots The Minifish was easier for Skipper to scoot around in with higher winds vs wrestling the larger Sunfish sail\ldots%
    } \\
    \midrule

    \orbtagthink{%
        The retriever conflates ``sunfish'' (a freshwater fish) with the \emph{Sunfish} sailboat brand---both content-level searches have failed.
        I should change strategy and anchor on the author instead.
        The question provides a unique biographical clue: the author was president of the \emph{American Institute of Human Paleontology}.
        Let me search for that role directly.%
    }
    \vspace{0.4em} \\

    \orbtagsearch{anthropologist-naturalist who became president of American Institute of Human Paleontology}
    \vspace{0.2em} \\

    \orbtaginfo{%
        \textbf{Doc 1 (Title: ``Gigantopithecus --- Wikipedia''):} Gigantopithecus has become popular in cryptozoology circles\ldots\\
        \textbf{Doc 2 (Title: ``The Immense Journey: An Imaginative Naturalist Explores the\ldots''):} \textbf{Anthropologist and naturalist Loren Eiseley} blends scientific knowledge and imaginative vision in this story of man. About the Author.\\
        \textbf{Doc 4 (Title: ``Ernest Chantre: French anthropologist, naturalist and archaeologist''):} Quick Facts \ldots French anthropologist, naturalist and archaeologist.%
    } \\
    \midrule

    \orbtagthink{%
        Doc~2 is the breakthrough: it names \textbf{Loren Eiseley} as ``anthropologist and naturalist'' \emph{and} mentions the title ``\textbf{The Immense Journey}''.
        I need to confirm this book was published in 1957.%
    }
    \vspace{0.4em} \\

    \orbtagsearch{Loren Eiseley 1957 essay collection}
    \vspace{0.2em} \\

    \orbtaginfo{%
        \textbf{Doc 4 (Title: ``The Immense Journey by Loren Eiseley --- The 5925th greatest book of\ldots''):} This page contains details about the book \textbf{The Immense Journey} by \textbf{Loren Eiseley published in 1957}.\\
        \textbf{Doc 5 (Title: ``The Immense Journey by Loren Eiseley (1957-08-12): Amazon.com\ldots''):} \ldots Eiseley compares and comments on the growing understanding of evolution by the scientific community---\textbf{until 1957 when he wrote this}.%
    } \\
    \midrule

    \orbtagthink{%
        The evidence converges: \textit{The Immense Journey} (1957), authored by anthropologist-naturalist \textit{Loren Eiseley}, matches every clue in the question---the year, the genre (nature essays), and the author's disciplinary identity.
        The institutional presidency clue could not be directly confirmed from retrieved documents, but the triple match on year, genre, and author profile is conclusive.%
    }\\
    \orbtagans{The Immense Journey} \\

    \bottomrule
    \end{tabularx}
    \caption{Case study of \orbit-4B during GRPO training on \orbit. Key evidence extracted from retrieved documents is \textbf{bolded}.}
    \label{tab:orbit-case-study-browsecomp}
\end{table*}
{\setlength{\tabcolsep}{6.5pt}
\renewcommand{\arraystretch}{0.8}
\fontsize{8.2}{12}\selectfont
\begin{longtable}{@{} p{\linewidth} @{}}
\caption{One training example from each of the 14 additional knowledge domains in \orbit. Colored spans mark the distinct factual clues embedded in the question; the verification rows confirm each clue with key supporting evidence (\textcolor{blue}{blue}).}
\label{tab:domain-examples} \\
\toprule
\endfirsthead
\toprule
\multicolumn{1}{c}{\emph{Table~\ref{tab:domain-examples} continued from previous page}}\\
\midrule
\endhead
\midrule
\multicolumn{1}{r}{\emph{continued on next page\ldots}}\\
\endfoot
\bottomrule
\endlastfoot

\rowcolor{odysseyrow} \multicolumn{1}{c}{\textbf{Domain: \emph{(2) Science \& Technology}}} \\ \midrule
\textbf{Question:} What term describes \Hblue{the mathematical concept first systematically
developed in a 1939 physics textbook} by \Horange{a Nobel laureate who independently predicted
antimatter}, where \Hpurple{angled brackets denote dual vector spaces and vertical bars separate
state vectors}, \Hgreen{revolutionizing the formalism of quantum measurement theory}? \\
\textbf{Answer: bra-ket notation} \\ \midrule
\rowcolor{odysseyrow} \multicolumn{1}{l}{{\textbf{Verification~/~Summary of Supporting Facts:}}} \\
\textcolor[HTML]{336600}{\cmark} \Hblue{\textbf{Origin:}} \emph{\textcolor{blue}{Paul Dirac} introduced bra-ket notation in his 1939 publication \emph{A New Notation for Quantum Mechanics}.} \\
\textcolor[HTML]{336600}{\cmark} \Horange{\textbf{Nobel Laureate:}} \emph{Dirac independently predicted \textcolor{blue}{antimatter (positrons)} through his equation, confirmed experimentally in 1932.} \\
\textcolor[HTML]{336600}{\cmark} \Hpurple{\textbf{Notation:}} \emph{Uses angled brackets $\langle$ (bras) and vertical bars $|$ (kets) to \textcolor{blue}{denote dual vector spaces and separate state vectors}.} \\
\textcolor[HTML]{336600}{\cmark} \Hgreen{\textbf{Impact:}} \emph{The notation \textcolor{blue}{revolutionized quantum measurement theory} by formalizing state vectors, operators, and inner products.} \\
\addlinespace[8pt]
\midrule

\rowcolor{odysseyrow} \multicolumn{1}{c}{\textbf{Domain: \emph{(3) Art}}} \\ \midrule
\textbf{Question:} In what year did the \Hblue{Cologne-based New Objectivity painter}, whose work
was \Horange{confiscated during the state-sponsored campaign against ``degenerate'' art} and who
\Hpurple{created psychologically charged portraits of women} while \Hgreen{married to an
avant-garde artist known for depictions of solitary male figures}, pass away after
\Hyellow{surviving wartime persecution and the destruction of her studio}? \\
\textbf{Answer: 1970} \\ \midrule
\rowcolor{odysseyrow} \multicolumn{1}{l}{{\textbf{Verification~/~Summary of Supporting Facts:}}} \\
\textcolor[HTML]{336600}{\cmark} \Hblue{\textbf{Painter:}} \emph{\textcolor{blue}{Marta Hegemann}, Cologne-based New Objectivity painter associated with the Cologne Progressives.} \\
\textcolor[HTML]{336600}{\cmark} \Horange{\textbf{Persecution:}} \emph{Her work was \textcolor{blue}{confiscated as ``degenerate art''} during the Nazi campaign.} \\
\textcolor[HTML]{336600}{\cmark} \Hpurple{\textbf{Style:}} \emph{She created \textcolor{blue}{psychologically charged portraits of women}.} \\
\textcolor[HTML]{336600}{\cmark} \Hgreen{\textbf{Spouse:}} \emph{Married to \textcolor{blue}{Anton R{\"a}derscheidt}, known for depictions of solitary male figures.} \\
\textcolor[HTML]{336600}{\cmark} \Hyellow{\textbf{Wartime:}} \emph{Endured \textcolor{blue}{wartime persecution and the destruction of her studio} during WWII; died in obscurity in 1970.} \\

\addlinespace[8pt]
\midrule

\rowcolor{odysseyrow} \multicolumn{1}{c}{\textbf{Domain: \emph{(4) History}}} \\ \midrule
\textbf{Question:} What was the total number of personnel demobilized following \Hblue{the largest
naval uprising in February 1946} that was \Horange{directly inspired by the trial of soldiers from
a colonial-era army unit}, which itself \Hpurple{became a catalyst for the final phase of a major
20th-century liberation struggle}? \\
\textbf{Answer: 476} \\ \midrule
\rowcolor{odysseyrow} \multicolumn{1}{l}{{\textbf{Verification~/~Summary of Supporting Facts:}}} \\
\textcolor[HTML]{336600}{\cmark} \Hblue{\textbf{Uprising:}} \emph{The \textcolor{blue}{Royal Indian Navy mutiny} (Feb.\ 1946) was the largest naval uprising, involving over 20,000 sailors.} \\
\textcolor[HTML]{336600}{\cmark} \Horange{\textbf{Trigger:}} \emph{Directly inspired by the \textcolor{blue}{trials of Indian National Army (INA) soldiers}, a colonial-era unit.} \\
\textcolor[HTML]{336600}{\cmark} \Hpurple{\textbf{Impact:}} \emph{Acted as a \textcolor{blue}{catalyst for India's independence} struggle, leading to British withdrawal in 1947.} \\
\textcolor[HTML]{336600}{\cmark} \textbf{Outcome:} \emph{Following the mutiny, \textcolor{blue}{476 sailors were dismissed} and were not reinstated post-independence.} \\

\addlinespace[8pt]
\midrule

\rowcolor{odysseyrow} \multicolumn{1}{c}{\textbf{Domain: \emph{(5) Sports}}} \\ \midrule
\textbf{Question:} In \Hblue{the Summer Olympics subject to a major geopolitical boycott}, which
men's track event awarded gold to \Horange{an athlete from a participating European nation} who
\Hpurple{had set a world record in a different middle-distance event earlier that same year at a
competition in a founding NATO country}, and whose \Hgreen{silver medalist represented a
Soviet-bloc nation that defied the boycott}? \\
\textbf{Answer: Men's 1500 metres} \\ \midrule
\rowcolor{odysseyrow} \multicolumn{1}{l}{{\textbf{Verification~/~Summary of Supporting Facts:}}} \\
\textcolor[HTML]{336600}{\cmark} \Hblue{\textbf{Olympics:}} \emph{The \textcolor{blue}{1980 Moscow Olympics} were boycotted by the United States and 65+ nations due to the Soviet invasion of Afghanistan.} \\
\textcolor[HTML]{336600}{\cmark} \Horange{\textbf{Gold Medalist:}} \emph{\textcolor{blue}{Sebastian Coe} of Great Britain won gold in the men's 1500m; Britain participated in the Games.} \\
\textcolor[HTML]{336600}{\cmark} \Hpurple{\textbf{World Record:}} \emph{Coe set the \textcolor{blue}{800m world record (1:42.33)} in Oslo, Norway (a founding NATO member), in July 1979.} \\
\textcolor[HTML]{336600}{\cmark} \Hgreen{\textbf{Silver Medalist:}} \emph{\textcolor{blue}{J{\"u}rgen Straub} represented East Germany, a Soviet-bloc nation that participated in the 1980 Games.} \\
\addlinespace[8pt]
\midrule

\rowcolor{odysseyrow} \multicolumn{1}{c}{\textbf{Domain: \emph{(6) Music}}} \\ \midrule
\textbf{Question:} During what year was \Hblue{the only presentation made for a specialized Grammy category honoring dance music productions}, awarded to \Horange{a female vocalist's anthem of resilience} recorded at \Hpurple{a studio previously used by a British rock band for their live
album featuring a cover of ``Twist and Shout,''} coinciding with \Hgreen{the ceremony where a singer-songwriter won Album of the Year}? \\
\textbf{Answer: 1980} \\ \midrule
\rowcolor{odysseyrow} \multicolumn{1}{l}{{\textbf{Verification~/~Summary of Supporting Facts:}}} \\
\textcolor[HTML]{336600}{\cmark} \Hblue{\textbf{Grammy Category:}} \emph{The \textcolor{blue}{Grammy Award for Best Disco Recording} was presented only once, at the 1980 ceremony.} \\
\textcolor[HTML]{336600}{\cmark} \Horange{\textbf{Winner:}} \emph{\textcolor{blue}{Gloria Gaynor's \emph{I Will Survive}} won the award; the song became a global anthem of resilience.} \\
\textcolor[HTML]{336600}{\cmark} \Hpurple{\textbf{Studio:}} \emph{Recorded at \textcolor{blue}{Atlantic Studios}, where The Beatles tracked their 1963 \emph{Twist and Shout} EP cover.} \\
\textcolor[HTML]{336600}{\cmark} \Hgreen{\textbf{Ceremony:}} \emph{\textcolor{blue}{Christopher Cross} won Album of the Year at the same 1980 Grammy Awards ceremony.} \\
\addlinespace[8pt]
\midrule

\rowcolor{odysseyrow} \multicolumn{1}{c}{\textbf{Domain: \emph{(7) Video Games}}} \\ \midrule
\textbf{Question:} What is the name of the \Hblue{summon creature associated with a mountain
location in a 1997 PlayStation RPG}, where this entity's \Horange{Japanese name derives from a
historical Korean military unit} and \Hpurple{appears in a spin-off title directed by}
\Hgreen{the developer who later created a 2006 action-RPG featuring a protagonist with amnesia}? \\
\textbf{Answer: Hiry\={u}} \\ \midrule
\rowcolor{odysseyrow} \multicolumn{1}{l}{{\textbf{Verification~/~Summary of Supporting Facts:}}} \\
\textcolor[HTML]{336600}{\cmark} \Hblue{\textbf{Origin:}} \emph{\textcolor{blue}{\emph{Final Fantasy Tactics} (1997)} includes the summon \emph{Hiry\={u}} associated with Mount Bervenia.} \\
\textcolor[HTML]{336600}{\cmark} \Horange{\textbf{Name:}} \emph{Hiry\={u}'s Japanese name derives from the \textcolor{blue}{historical Korean Hwarang military unit}.} \\
\textcolor[HTML]{336600}{\cmark} \Hpurple{\textbf{Spin-off:}} \emph{Hiry\={u} \textcolor{blue}{appears in \emph{Final Fantasy XII: Revenant Wings}}, directed by Yasumi Matsuno.} \\
\textcolor[HTML]{336600}{\cmark} \Hgreen{\textbf{Director:}} \emph{\textcolor{blue}{Yasumi Matsuno} later directed \emph{Final Fantasy XII} (2006), featuring protagonist Vaan with fragmented memories.} \\
\addlinespace[8pt]
\midrule

\rowcolor{odysseyrow} \multicolumn{1}{c}{\textbf{Domain: \emph{(8) Geography}}} \\ \midrule
\textbf{Question:} What is the name of the city \Hblue{designated as World Book Capital for 2023},
which \Horange{became the first in its geographical region to receive this honor}, known for
\Hpurple{hosting an annual international book fair that attracts publishers from French-speaking
countries}, and whose selection was \Hgreen{announced by UNESCO in a city that shares its name
with a famous aviation pioneer}? \\
\textbf{Answer: Accra} \\ \midrule
\rowcolor{odysseyrow} \multicolumn{1}{l}{{\textbf{Verification~/~Summary of Supporting Facts:}}} \\
\textcolor[HTML]{336600}{\cmark} \Hblue{\textbf{Designation:}} \emph{\textcolor{blue}{Accra} was designated as the UNESCO World Book Capital for 2023.} \\
\textcolor[HTML]{336600}{\cmark} \Horange{\textbf{Region:}} \emph{First city in \textcolor{blue}{West Africa} (third in Sub-Saharan Africa) to receive the title.} \\
\textcolor[HTML]{336600}{\cmark} \Hpurple{\textbf{Book Fair:}} \emph{Accra hosts the \textcolor{blue}{Ghana International Book Fair}, attracting publishers from French-speaking countries.} \\
\textcolor[HTML]{336600}{\cmark} \Hgreen{\textbf{Announcement:}} \emph{Selection announced by UNESCO in \textcolor{blue}{Paris}, the city named after the aviation pioneer Alberto Santos-Dumont's adopted home.} \\
\addlinespace[8pt]
\midrule

\rowcolor{odysseyrow} \multicolumn{1}{c}{\textbf{Domain: \emph{(9) Politics}}} \\ \midrule
\textbf{Question:} At which major political convention did \Hblue{a coalition encompassing both
moderate and conservative factions} experience \Horange{a platform dispute over foreign policy morality}, resulting in the \Hpurple{narrow defeat (by 60~votes) of a challenger who had advocated for that plank}, during \Hgreen{the last election cycle where this broad-coalition party failed to retain the presidency}? \\
\textbf{Answer: 1976 Republican National Convention} \\ \midrule
\rowcolor{odysseyrow} \multicolumn{1}{l}{{\textbf{Verification~/~Summary of Supporting Facts:}}} \\
\textcolor[HTML]{336600}{\cmark} \Hblue{\textbf{Factions:}} \emph{Conflict between \textcolor{blue}{moderate President Gerald Ford and conservative challenger Ronald Reagan} at the 1976 RNC.} \\
\textcolor[HTML]{336600}{\cmark} \Horange{\textbf{Platform Dispute:}} \emph{Reagan proposed a \textcolor{blue}{``foreign policy morality'' plank} criticising Ford's d{\'e}tente policy.} \\
\textcolor[HTML]{336600}{\cmark} \Hpurple{\textbf{Vote:}} \emph{Reagan's morality plank was \textcolor{blue}{narrowly defeated by approximately 60~votes} at the convention.} \\
\textcolor[HTML]{336600}{\cmark} \Hgreen{\textbf{Election Outcome:}} \emph{The 1976 cycle ended with the Republican Party \textcolor{blue}{losing the presidency to Jimmy Carter}.} \\
\addlinespace[8pt]
\midrule

\rowcolor{odysseyrow} \multicolumn{1}{c}{\textbf{Domain: \emph{(10) Medicine}}} \\ \midrule
\textbf{Question:} What is the specific term for the phenomenon where \Hblue{certain small
molecules form colloidal aggregates in aqueous buffer}, leading to \Horange{false positive results
in biochemical assays} that measure activity through \Hpurple{optical detection methods} commonly
employed in \Hgreen{large-scale compound evaluation}? \\
\textbf{Answer: colloidal aggregation} \\ \midrule
\rowcolor{odysseyrow} \multicolumn{1}{l}{{\textbf{Verification~/~Summary of Supporting Facts:}}} \\
\textcolor[HTML]{336600}{\cmark} \Hblue{\textbf{Mechanism:}} \emph{Small molecules \textcolor{blue}{form colloidal aggregates in aqueous buffer}, sequestering assay proteins.} \\
\textcolor[HTML]{336600}{\cmark} \Horange{\textbf{Consequence:}} \emph{Aggregates produce \textcolor{blue}{false positive results in biochemical assays} by non-specifically inhibiting enzymes.} \\
\textcolor[HTML]{336600}{\cmark} \Hpurple{\textbf{Detection:}} \emph{Commonly detected via \textcolor{blue}{optical methods} (absorbance/turbidity) in high-throughput screening.} \\
\textcolor[HTML]{336600}{\cmark} \Hgreen{\textbf{Context:}} \emph{A major artifact in \textcolor{blue}{large-scale compound evaluation} (e.g., drug discovery HTS campaigns).} \\
\addlinespace[8pt]
\midrule

\rowcolor{odysseyrow} \multicolumn{1}{c}{\textbf{Domain: \emph{(11) Finance}}} \\ \midrule
\textbf{Question:} Which financial valuation component \Hblue{equals the difference between a
stock's market price and its no-growth value}, \Horange{requires forecasting of sustainable
competitive advantages}, \Hpurple{incorporates dividend payout ratios in its calculation},
\Hgreen{is inversely related to the cost of equity capital}, and \Hyellow{becomes negative when a
firm's return on equity falls below its required return}? \\
\textbf{Answer: PVGO} \\ \midrule
\rowcolor{odysseyrow} \multicolumn{1}{l}{{\textbf{Verification~/~Summary of Supporting Facts:}}} \\
\textcolor[HTML]{336600}{\cmark} \Hblue{\textbf{Definition:}} \emph{\textcolor{blue}{PVGO} (Present Value of Growth Opportunities) $=$ Market Price $-$ No-Growth Value.} \\
\textcolor[HTML]{336600}{\cmark} \Horange{\textbf{Input:}} \emph{Requires \textcolor{blue}{forecasting sustainable competitive advantages} to estimate future excess returns.} \\
\textcolor[HTML]{336600}{\cmark} \Hpurple{\textbf{Formula:}} \emph{Incorporates \textcolor{blue}{dividend payout ratios} in the Gordon Growth Model decomposition.} \\
\textcolor[HTML]{336600}{\cmark} \Hgreen{\textbf{Relationship:}} \emph{\textcolor{blue}{Inversely related to the cost of equity capital}; a higher discount rate compresses PVGO.} \\
\textcolor[HTML]{336600}{\cmark} \Hyellow{\textbf{Sign:}} \emph{PVGO turns \textcolor{blue}{negative when ROE $<$ required return}, signalling value-destroying reinvestment.} \\
\addlinespace[8pt]
\midrule

\rowcolor{odysseyrow} \multicolumn{1}{c}{\textbf{Domain: \emph{(12) Law}}} \\ \midrule
\textbf{Question:} What is the name of the \Hblue{landmark US Supreme Court case that upheld a state's authority to sterilize individuals deemed ``unfit,''} where the majority opinion was
\Horange{authored by a justice who served concurrently with} and \Hpurple{frequently dissented
from another justice famous for his ``clear and present danger'' test}? \\
\textbf{Answer: \emph{Buck v.\ Bell}} \\ \midrule
\rowcolor{odysseyrow} \multicolumn{1}{l}{{\textbf{Verification~/~Summary of Supporting Facts:}}} \\
\textcolor[HTML]{336600}{\cmark} \Hblue{\textbf{Case:}} \emph{\textcolor{blue}{\emph{Buck v.\ Bell}} upheld Virginia's authority to sterilize Carrie Buck under the 1924 Eugenical Sterilization Act.} \\
\textcolor[HTML]{336600}{\cmark} \Horange{\textbf{Author:}} \emph{Majority written by \textcolor{blue}{Justice Oliver Wendell Holmes Jr.}, declared ``Three generations of imbeciles are enough.''} \\
\textcolor[HTML]{336600}{\cmark} \Hpurple{\textbf{Colleague:}} \emph{Holmes served alongside \textcolor{blue}{Justice Louis Brandeis}, co-author of the ``clear and present danger'' test (\emph{Schenck v.\ United States}, 1919).} \\
\textcolor[HTML]{336600}{\cmark} \textbf{Legacy:} \emph{The ruling remains \textcolor{blue}{unoverturned}, enabling over 60,000 forced sterilizations across the United States.} \\
\addlinespace[8pt]
\midrule

\rowcolor{odysseyrow} \multicolumn{1}{c}{\textbf{Domain: \emph{(13) Mathematics}}} \\ \midrule
\textbf{Question:} What is the name of the \Hblue{measure of statistical dispersion often used for
economic inequality} that can be \Horange{calculated using a method based on ranking and indexing},
\Hpurple{developed by an Italian statistician} who also \Hgreen{made significant contributions to
the study of variability and demographic analysis}? \\
\textbf{Answer: Gini coefficient} \\ \midrule
\rowcolor{odysseyrow} \multicolumn{1}{l}{{\textbf{Verification~/~Summary of Supporting Facts:}}} \\
\textcolor[HTML]{336600}{\cmark} \Hblue{\textbf{Measure:}} \emph{The \textcolor{blue}{Gini coefficient} is the standard measure of statistical dispersion for economic inequality.} \\
\textcolor[HTML]{336600}{\cmark} \Horange{\textbf{Calculation:}} \emph{Computed via the \textcolor{blue}{Lorenz curve}, which involves ranking income or wealth from lowest to highest.} \\
\textcolor[HTML]{336600}{\cmark} \Hpurple{\textbf{Author:}} \emph{Developed by \textcolor{blue}{Corrado Gini} (1884--1965), Italian statistician.} \\
\textcolor[HTML]{336600}{\cmark} \Hgreen{\textbf{Contributions:}} \emph{Gini made \textcolor{blue}{foundational contributions to variability theory and demography}.} \\
\addlinespace[8pt]
\midrule

\rowcolor{odysseyrow} \multicolumn{1}{c}{\textbf{Domain: \emph{(14) Puzzles}}} \\ \midrule
\textbf{Question:} What is the title of the \Hblue{1995 anti-gravity racing game} developed and
published by \Horange{the same studio that released a puzzle game featuring suicidal green-haired
creatures}, which became a \Hpurple{European launch title for a 32-bit console} and \Hgreen{featured
licensed tracks from electronic musicians including a duo known for a tree-themed debut album}? \\
\textbf{Answer: Wipeout} \\ \midrule
\rowcolor{odysseyrow} \multicolumn{1}{l}{{\textbf{Verification~/~Summary of Supporting Facts:}}} \\
\textcolor[HTML]{336600}{\cmark} \Hblue{\textbf{Game:}} \emph{\textcolor{blue}{Wipeout} (1995) was an anti-gravity racing game developed by Psygnosis.} \\
\textcolor[HTML]{336600}{\cmark} \Horange{\textbf{Studio:}} \emph{\textcolor{blue}{Psygnosis} also published \emph{Lemmings} (1991), the puzzle game featuring green-haired suicidal creatures.} \\
\textcolor[HTML]{336600}{\cmark} \Hpurple{\textbf{Console:}} \emph{Wipeout was a \textcolor{blue}{European launch title for the PlayStation} (32-bit) in 1995.} \\
\textcolor[HTML]{336600}{\cmark} \Hgreen{\textbf{Soundtrack:}} \emph{Featured \textcolor{blue}{licensed electronic tracks} from Orbital and The Chemical Brothers.} \\
\addlinespace[8pt]
\midrule

\rowcolor{odysseyrow} \multicolumn{1}{c}{\textbf{Domain: \emph{(15) Code}}} \\ \midrule
\textbf{Question:} What is the name of the \Hblue{algebraic structure formed by the equivalence
classes of sentences} under the relation that holds when \Horange{two sentences have the same
truth value in all possible interpretations}, as introduced by \Hpurple{a logician who also
formulated a semantic definition of truth for formal languages} and \Hgreen{collaborated on a
paradox involving the decomposition of spheres}? \\
\textbf{Answer: Lindenbaum algebra} \\ \midrule
\rowcolor{odysseyrow} \multicolumn{1}{l}{{\textbf{Verification~/~Summary of Supporting Facts:}}} \\
\textcolor[HTML]{336600}{\cmark} \Hblue{\textbf{Structure:}} \emph{The \textcolor{blue}{Lindenbaum algebra} consists of equivalence classes of sentences under logical equivalence.} \\
\textcolor[HTML]{336600}{\cmark} \Horange{\textbf{Equivalence:}} \emph{The relation holds when two sentences share the \textcolor{blue}{same truth value in all possible interpretations}.} \\
\textcolor[HTML]{336600}{\cmark} \Hpurple{\textbf{Author:}} \emph{\textcolor{blue}{Alfred Tarski} formulated the semantic definition of truth for formal languages and co-named the structure.} \\
\textcolor[HTML]{336600}{\cmark} \Hgreen{\textbf{Collaboration:}} \emph{Tarski collaborated with Stefan Banach on the \textcolor{blue}{Banach--Tarski paradox} (decomposition of a sphere into two identical copies).} \\

\end{longtable}
}

\clearpage
\small
\setlength{\tabcolsep}{3pt}

\begin{longtable}{p{\linewidth}}
\caption{Top 100 categories per each domain in the \orbit dataset generated using OpenAI's Deep Research tool.}\label{tab:openai-categories} \\
\toprule
\rowcolor{white} \textbf{List of up to 100 Wikipedia Categories} (generated using OpenAI's Deep Research) \\
\midrule
\endfirsthead

\toprule
\textbf{List of up to 100 Wikipedia Categories} (continued) \\
\midrule
\endhead

\midrule
\multicolumn{1}{r}{\emph{Continued on next page}}  \\
\midrule
\endfoot

\bottomrule
\endlastfoot

\rowcolor{odysseyrow} \multicolumn{1}{c}{\textbf{\emph{(1) TV Shows \& Movies}}} \\
\addlinespace[4pt]
American crime drama television series; British science fiction television series; Japanese drama television series; Anime television series; Sitcoms; Reality television series; Game shows; Miniseries; Documentary television series; Animated films; Action films; Comedy films; Horror films; Science fiction films; Documentary films; Silent films; Black-and-white films; Direct-to-video films; Short films; American independent films; Spaghetti Western films; Film noir; Kaiju; Blaxploitation films; Animated feature films; 1970s American television series; Hindi-language films; French New Wave; South Korean films; Musical films; Films about outer space; Films about immigration to the United States; Films about poverty; Films about educators; Films about viral outbreaks; Films about artificial intelligence; Films about time travel; Films based on actual events; Lost films; Unfinished films; Cancelled films; Directorial debut films; Film series; Film remakes; Hong Kong action films; Direct-to-video horror films; British comedy films; Indian horror films; French animated films; Anime films; Soap operas; Telenovelas; Netflix original programming; HBO original programming; Anthology television series; Serial killer films; Superhero films; Monster movies; Spy films; Biographical films; Fantasy films; War films; Disaster films; Adventure films; Thriller films; Psychological thriller films; Slasher films; Dystopian films; Post-apocalyptic films; Biblical films; Christmas films; Best Picture Academy Award winners; Film festivals; Television spin-offs; Unaired television pilots; Propaganda films; Crossover films; Video nasties; 3D films; IMAX films; Peplum films; Neo-noir films; Samurai films; Martial arts films; African films; Mexican films; Iranian films; Films based on video games; Television series based on films; Propaganda films; Crossover films; Film posters by genre; Remake television series; Unaired television pilots; Video nasties; 3D films; IMAX films. \\  

\addlinespace[8pt]
\midrule

\rowcolor{odysseyrow} \multicolumn{1}{c}{\textbf{\emph{(2) Science \& Technology}}} \\ \addlinespace[4pt]
Physics; Chemistry; Biology; Mathematics; Astronomy; Geology; Computer science; Engineering; Materials science; Nanotechnology; Robotics; Artificial intelligence; Machine learning; Quantum mechanics; Quantum field theory; General relativity; Thermodynamics; Particle physics; Nuclear physics; Optics; Astrophysics; Spaceflight; Satellites; Rocketry; Renewable energy; Climate change; Meteorology; Oceanography; Volcanology; Ecology; Evolutionary biology; Genetics; Bioinformatics; Neuroscience; Molecular biology; Virology; Biomedical engineering; Biotechnology; Genetic engineering; Cryonics; Telescopes; Microscopy; Chemical elements; Organic compounds; Polymers; Crystallography; Electronics; Semiconductors; Internet; Computer security; Operating systems; Programming languages; Open-source software; Computer networks; Supercomputers; Distributed computing; Cloud computing; Databases; Algorithms; Graph theory; Number theory; Combinatorics; Topology; Algebra; Geometry; Differential equations; Computational biology; Astrobiology; Exoplanets; Dark matter; String theory; Complex systems; Chaos theory; Computer vision; Natural language processing; Human–computer interaction; Embedded systems; Quantum computing; Virtual reality; Augmented reality; Blockchains; Mobile phones; Telecommunications; Software engineering; Programming paradigms; Data structures; Artificial life; Cryptography; Information theory; Game theory; Automation; Aerospace engineering; Mechanical engineering; Electrical engineering; Civil engineering; Chemical engineering; Computational chemistry; Systems biology; Computational complexity theory. \\  \addlinespace[8pt] \midrule

\rowcolor{odysseyrow} \multicolumn{1}{c}{\textbf{\emph{(3) Art}}} \\ \addlinespace[4pt]
Painting; Sculpture; Photography; Architecture; Drawing; Printmaking; Ceramics; Textile arts; Glass art; Graphic design; Industrial design; Fashion design; Calligraphy; Comics; Theatre; Opera; Dance; Ballet; Circus; Magic (illusion); Performance art; Street art; Public art; Digital art; Contemporary art; Modern art; Renaissance art; Medieval art; Baroque art; Impressionism; Post-Impressionism; Surrealism; Cubism; Futurism; Dada; Abstract art; Abstract expressionism; Pop art; Art Nouveau; Gothic architecture; Baroque architecture; Modernist architecture; Skyscrapers; Video art; Installation art; Conceptual art; Murals; Portrait painting; Landscape art; Still life paintings; Mosaics; Poetry; Novels; Short stories; Plays; Science fiction novels; Fantasy novels; Detective novels; Children's literature; African art; Islamic art; Pre-Columbian art; Chinese art; Indian art; Japanese art; Indigenous Australian art; Mexican art; Native American art; Art history; Byzantine art; Christian art; Stagecraft; Choreography; Typography; Heraldry; Vexillology; Cartography; History painting; Nude art; Religious art; Decorative arts; Folk art; Outsider art; Manga; Musicals; Costume design; Black-and-white photography; Portrait photography; Renaissance architecture; Neoclassical architecture; Landscape architecture; French art; American art; Latin American art; Mime; Puppetry; Vaudeville; Improvisational theatre. \\ \addlinespace[8pt] \midrule

\rowcolor{odysseyrow} \multicolumn{1}{c}{\textbf{\emph{(4) History}}} \\ \addlinespace[4pt]
Prehistory; Ancient history; Classical antiquity; Middle Ages; Renaissance; Age of Enlightenment; Age of Discovery; Industrial Revolution; French Revolution; American Revolutionary War; Napoleonic Wars; World War I; World War II; Cold War; Byzantine Empire; Ottoman Empire; Roman Empire; Ancient Egypt; Ancient Greece; Mesopotamia; Ming dynasty; Victorian era; Meiji period; Maya civilization; History of Europe; History of Asia; History of Africa; History of North America; History of the United States; History of China; History of India; History of science; History of medicine; Economic history; Military history; History of technology; History of literature; Maritime history; Historiography; Archaeology; Revolutions; Rebellions; Colonialism; Imperialism; Slavery; Crusades; European colonization of the Americas; Scramble for Africa; Exploration; Silk Road; Atlantic slave trade; Battles; Sieges; 19th century; 20th century; Bronze Age; Iron Age; Neolithic; Scientific Revolution; Space Age; Information Age; Space Race; Human migrations; Paleontology; World Heritage Sites; American Civil War; Russian Revolution; Spanish Civil War; The Holocaust; Genocide; Civil rights movement; Women's suffrage; Decolonization; Great Depression; Famines; Plagues; Age of Discovery; Achaemenid Empire; Mongol Empire; British Empire; Holy Roman Empire; Soviet Union; Cultural Revolution; Apartheid; American Old West; Sengoku period; History of printing; Illuminated manuscripts; Feudalism; History of New York City; History of the Catholic Church; History of Islam; History of Christianity; History of science fiction; History of mathematics; History of computing; Ancient warfare; Nuclear warfare. \\ \addlinespace[8pt] \midrule

\rowcolor{odysseyrow} \multicolumn{1}{c}{\textbf{\emph{(5) Sports}}} \\ \addlinespace[4pt]
Association football; Basketball; Cricket; Baseball; American football; Rugby union; Rugby league; Tennis; Golf; Athletics (sport); Gymnastics; Swimming; Skiing; Ice hockey; Figure skating; Boxing; Martial arts; Mixed martial arts; Wrestling; Weightlifting; Cycling; Motorsport; Formula One; Chess; Extreme sports; Team sports; Individual sports; Combat sports; Water sports; Winter sports; Parasports; Women's sports; Youth sport; Olympic Games; Paralympic Games; Commonwealth Games; FIFA World Cup; UEFA European Championship; UEFA Champions League; Cricket World Cup; Super Bowl; World Series; Sport in India; Sport in the United States; College sports; High school sports; Olympic sports; Sports equipment; Sports venues; Doping in sports; Football hooliganism; Match fixing; Sports injuries; Mountaineering; Rock climbing; Underwater sports; Surfing; Skateboarding; Snowboarding; Shooting sports; Archery; Equestrian sports; Horse racing; Sumo; Kabaddi; Sport in Australia; International sports competitions; World championships; Motorcycle racing; Endurance sports; Marathons; Triathlon; Parkour; Professional wrestling; Orienteering; Table tennis; Badminton; Volleyball; Handball; Cue sports; Judo; Asian Games; Pan American Games; X Games; Winter Olympic Games; Summer Olympic Games; African Games; Speed skating; Karate; Taekwondo; Kickboxing; Esports. \\ \addlinespace[8pt] \midrule

\rowcolor{odysseyrow} \multicolumn{1}{c}{\textbf{\emph{(6) Music}}} \\ \addlinespace[4pt]
Classical music; Rock music; Pop music; Hip hop music; Jazz; Blues; Country music; Electronic music; Heavy metal music; Punk rock; Folk music; Reggae; Soul music; Gospel music; Rhythm and blues; Funk; Latin music; K-pop; Hindustani music; Carnatic music; Japanese music; Brazilian music; Nigerian music; Baroque music; Classical period (music); Romantic music; 20th-century classical music; Contemporary classical music; Medieval music; Renaissance music; World music; Concept albums; Live albums; Soundtrack albums; Compilation albums; Debut albums; Instrumental music; A cappella; Choral music; Symphonies; Music theory; Musicology; Musical instruments; String instruments; Wind instruments; Percussion instruments; Keyboard instruments; Electronic musical instruments; Music festivals; Eurovision Song Contest; Music videos; Film scores; House music; Techno; Ambient music; Trance music; Drum and bass; Heavy metal genres; Progressive rock; Alternative rock; Hardcore punk; Doom metal; Death metal; Black metal; Thrash metal; Industrial music; New wave music; Disco; Salsa music; Tango music; Flamenco; Bhangra (music); Afrobeat; Highlife; J-pop; Mandopop; Barbershop music; Contemporary Christian music; Oratorios; Concertos; Operettas; National anthems; March music; Video game music; Chiptune; Music industry; Sound recording; Audio engineering; DJing; Hip hop genres; Trap music; Reggaeton; Karaoke; One-hit wonders; Christmas music; Patriotic songs; Protest songs; Ballads; Lullabies; Remix albums. \\ \addlinespace[8pt] \midrule

\rowcolor{odysseyrow} \multicolumn{1}{c}{\textbf{\emph{(7) Video Games}}} \\ \addlinespace[4pt]
Action games; Adventure games; Role-playing video games; Simulation video games; Strategy video games; Puzzle video games; Platform games; Racing video games; Sports video games; First-person shooters; Third-person shooters; Survival horror video games; Stealth video games; Fighting games; Beat 'em up games; Visual novels; Massively multiplayer online games; Mobile games; Browser games; Arcade games; Indie video games; Open-source video games; Freeware games; Nintendo Switch games; PlayStation 4 games; Japanese video games; American video games; Indian video games; 2020 video games; Video games based on films; Video games based on novels; World War II video games; Educational video games; Christian video games; LGBTQ-related video games; Horror video games; Science fiction video games; Fantasy video games; Cyberpunk video games; Dystopian video games; Post-apocalyptic video games; Alternate history video games; Crossover video games; Game engines; Game design; Video game franchises; Mario video games; Pokémon video games; Sonic the Hedgehog video games; Final Fantasy video games; Resident Evil video games; Call of Duty video games; The Sims games; Tomb Raider games; Street Fighter games; Mortal Kombat video games; The Legend of Zelda games; Grand Theft Auto games; Star Wars video games; Star Trek games; DC Comics video games; Marvel Comics video games; Cross-platform video games; Roguelike video games; Metroidvania games; 4X video games; Tower defense video games; Dating sims; Incremental games; Virtual reality games; Augmented reality games; Text adventure games; Interactive fiction; Windows games; MacOS games; Linux games; Game Boy games; PlayStation 2 games; Xbox One games; Video game controversies; Video games based on Marvel Comics; Video games based on DC Comics; Esports games; Cancelled video games; Dōjin games; Erotic video games; Multiplayer online battle arena games; Battle royale games; Open world video games; Action-adventure games; Shoot 'em up games; Rhythm games; Music video games; Video game compilations; Game jam video games; Video game mods; Unreal Engine games; Unity (game engine) games. \\ \addlinespace[8pt] \midrule

\rowcolor{odysseyrow} \multicolumn{1}{c}{\textbf{\emph{(8) Geography}}} \\ \addlinespace[4pt]
Geography of Africa; Geography of Asia; Geography of Europe; Geography of North America; Geography of South America; Geography of Oceania; Geography of Antarctica; Mountains; Rivers; Lakes; Islands; Deserts; Forests; Volcanoes; Caves; Canyons; Waterfalls; Glaciers; Oceans; Seas; National parks; Protected areas; Ghost towns; Megacities; Capital cities; Ports and harbours; Bridges; Tunnels; Mountain ranges; Seamounts; Oceanic trenches; Atolls; Coral reefs; Geographic poles; Extreme points of Earth; Maps; Earthquakes; Tsunamis; Tropical cyclones; Volcanic eruptions; Impact craters; Middle East; Scandinavia; Latin America; Arctic; Sahara; Himalayas; Alps; Andes; Amazon basin; Great Plains; Siberia; Pacific Ocean; Atlantic Ocean; Indian Ocean; Arctic Ocean; Southern Ocean; Mediterranean Sea; Caribbean Sea; Red Sea; Bays; Peninsulas; Valleys; Deltas; Swamps; Steppes; Plateaus; Oases; Ramsar sites; Urban planning; Time zones; Equator; Tropics; Reservoirs; Dams; Straits; Isthmuses; Beaches; Sand dunes; Mountain passes; Geysers; Countries in Africa; States of the United States; Metropolitan areas; Volcanic islands; Highlands; Rainforests; Tundra; Savannas; Monsoons; Climate zones; Permafrost; Settlements; Villages; Towns; Continents; Borders; Coasts; Transcontinental countries; Natural disasters. \\ \addlinespace[8pt] \midrule

\rowcolor{odysseyrow} \multicolumn{1}{c}{\textbf{\emph{(9) Politics}}} \\ \addlinespace[4pt]
Communism; Socialism; Capitalism; Anarchism; Conservatism; Liberalism; Fascism; Democracy; Monarchy; Republics; Parliamentary systems; Totalitarianism; Populism; Nationalism; Colonialism; Imperialism; Globalization; Human rights; Constitutions; Legislatures; Elections; Referendums; Political parties; Political ideologies; Revolutions; Coups d'état; Propaganda; Censorship; Political scandals; Corruption; Authoritarianism; Militarism; Separatism; Federalism; Unitary states; Dictatorships; United Nations; NATO; Non-governmental organizations; International relations; Diplomacy; Treaties; War; Peace movements; Disarmament; Genocides; Electoral systems; Public policy; Constitutional courts; Political campaigns; Lobbying; Political violence; Terrorism; Ideological conflicts; Civil wars; Social movements; Civil rights; Women's rights; LGBT rights; Environmentalism; Climate politics; Immigration; Gun politics; Drug policy; Healthcare reform; Education policy; Tax policy; City politics; Local government; Federal governments; Politics of the United States; Politics of the United Kingdom; Politics of India; Politics of France; Politics of the European Union; Politics of the United Nations; Elections by country; Political parties by country; Revolutions by country; Human rights by country; Military coups by country; Political corruption by country; Foreign relations by country; Populism by country; Nationalism by country; Cold War; War on Terror; Arab Spring; Democratization; Electoral fraud; Gerrymandering; Propaganda by country; Public opinion; Political communication; Freedom of speech; Campaign finance. \\ \addlinespace[8pt] \midrule

\rowcolor{odysseyrow} \multicolumn{1}{c}{\textbf{\emph{(10) Medicine}}} \\ \addlinespace[4pt]
Diseases; Infectious diseases; Genetic disorders; Cardiovascular diseases; Cancers; Neurological disorders; Mental disorders; Respiratory diseases; Skin conditions; Pandemics; Rare diseases; Pediatrics; Geriatrics; Oncology; Cardiology; Neurology; Psychiatry; Endocrinology; Immunology; Epidemiology; Public health; Preventive medicine; Occupational safety and health; Pharmacology; Clinical trials; Vaccines; Antibiotics; Medical devices; Prosthetics; Medical imaging; Diagnostics; First aid; Emergency medicine; Surgery; Anesthesiology; Transplantation medicine; Medical genetics; Pathology; Radiology; Dermatology; Ophthalmology; Obstetrics and gynecology; Orthopedics; Sports medicine; Dentistry; Veterinary medicine; Alternative medicine; Traditional medicine; Herbalism; Homeopathy; Chiropractic; Medical ethics; History of medicine; Medical education; Hospitals; Healthcare in the United States; Healthcare in the United Kingdom; Healthcare in India; Health law; Medical research; Evidence-based medicine; Global health; Nutrition; Occupational therapy; Physical therapy; Rehabilitation medicine; Pharmaceuticals; Drug discovery; Medical imaging; Telemedicine; Genomics; Stem cell research; Bioethics; Clinical psychology; Disease outbreaks; Quarantine; Tropical diseases; Neglected diseases; Zoonotic diseases; Endemic diseases; Chronic diseases; Terminal illnesses; Palliative care; Mortality; Surgery by specialty; Minimally invasive surgery; Robotic surgery; Plastic surgery; Medical imaging; Radiotherapy; Chemotherapy; Pharmacology; Over-the-counter drugs; Prescription drugs; Controlled substances; Addiction medicine; Outbreaks. \\ \addlinespace[8pt] \midrule

\rowcolor{odysseyrow} \multicolumn{1}{c}{\textbf{\emph{(11) Finance}}} \\ \addlinespace[4pt]
Finance; Economics; Macroeconomics; Microeconomics; Banking; Accounting; Personal finance; Corporate finance; Public finance; International finance; Financial markets; Stock market; Bond market; Foreign exchange market; Derivatives (finance); Commodities; Cryptocurrencies; Insurance; Real estate; Investing; Investment funds; Hedge funds; Private equity; Venture capital; Mutual funds; Index funds; Exchange-traded funds; Retirement plans; Pension funds; Wealth management; Asset management; Risk management; Financial risk; Credit; Loans; Mortgages; Interest rates; Inflation; Deflation; Central banks; Monetary policy; Fiscal policy; Economic growth; Recessions; Financial crises; Stock market crashes; Depressions (economics); Economic bubbles; Hyperinflation; International trade; Balance of payments; Exchange rates; Budget deficits; Public debt; Taxation; Tariffs; Economic history; Financial history; Economic systems; Capitalism; Socialism; Mixed economies; Market economies; Planned economies; Shadow economy; Underground economy; Financial regulation; Financial law; Bank regulation; Anti-money laundering; Financial technology; Digital currencies; Mobile payments; Online banking; High-frequency trading; Stock exchanges; Derivatives trading; Commodity markets; Real estate investment trusts; Housing market; Insurance companies; Reinsurance; Actuarial science; Corporate governance; Shareholder activism; Financial scandals; Insider trading; Accounting scandals; Pyramid schemes; Fraud; Ponzi schemes; Financial literacy; Personal budgeting; Philanthropy; Microfinance \\ \addlinespace[8pt] \midrule

\rowcolor{odysseyrow} \multicolumn{1}{c}{\textbf{\emph{(12) Law}}} \\ \addlinespace[4pt]
Criminal law; Civil law (legal system); Common law; Constitutional law; Administrative law; International law; Human rights law; Labour law; Environmental law; Intellectual property law; Patent law; Copyright law; Trademark law; Contract law; Tort law; Property law; Family law; Inheritance law; Tax law; Bankruptcy law; Immigration law; Criminal justice; Law enforcement; Evidence law; Legal history; Trials; Case law; Law of the United States; Law of the United Kingdom; Law of India; Sharia; Canon law; Roman law; Jurisprudence; Legal doctrines; Legal ethics; Law reform; War crimes; International criminal law; International humanitarian law; Crimes; Punishments; Prisons; Execution methods; Capital punishment; European Court of Human Rights cases; United States Supreme Court cases; International Court of Justice cases; Landmark court decisions; Arbitration; Mediation; Lawsuits; Class action lawsuits; Corruption; Money laundering; Counterterrorism; Cybercrime; Organized crime; Criminology; Forensic science; Fraud; Gun laws; Censorship; Drug control law; Martial law; Anti-discrimination law; Privacy law; Internet law; Space law; Law of the sea; Banking law; Insurance law; Competition law; Commercial law; Admiralty law; Election law; Natural law; Military law; Obscenity law; Same-sex marriage law; Abortion law; Blue laws; Legal codes; Constitutions; Supreme courts; Courts by country; Police misconduct; Wrongful convictions; Freedom of expression law; Blasphemy law; Indigenous law; Religious law; Customary law; Legal fictions; Extradition; International criminal tribunals; Arms control; International sanctions; Geneva Conventions; Nuremberg trials \\ \addlinespace[8pt] \midrule

\rowcolor{odysseyrow} \multicolumn{1}{c}{\textbf{\emph{(13) Mathematics}}} \\ \addlinespace[4pt]
Algebra; Mathematical analysis; Applied mathematics; Arithmetic; Calculus; Combinatorics; Computational mathematics; Discrete mathematics; Dynamical systems; Elementary mathematics; Experimental mathematics; Foundations of mathematics; Game theory; Geometry; Graph theory; Mathematical logic; Number theory; Order theory; Recreational mathematics; Topology; Probability; Statistics; Abstract algebra; Linear algebra; Prime numbers; Euclidean geometry; Non-Euclidean geometry; Algebraic geometry; Differential geometry; Algebraic topology; Mathematical physics; Chaos theory; Set theory; Category theory; Differential equations; Partial differential equations; History of mathematics; Philosophy of mathematics; Mathematics education; Women in mathematics; Mathematicians; Mathematics competitions; Mathematics organizations; Mathematics awards; Mathematical games; Theoretical computer science; Information theory; Fractals; Numbers; Probability distributions; Mathematical notation; Mathematical theorems; Mathematical proofs; Pseudomathematics; Mathematical tools; Mathematics and art; Mathematics and culture; Mathematical constants; Mathematical concepts; Polygons; Polyhedra; Curves; Surfaces; Polynomials; Special functions; Equations; Inequalities (mathematics); Temporal paradoxes; Philosophical paradoxes; Undecidable problems; NP-complete problems; Computational complexity theory; Combinatorial optimization; Linear programming; Coding theory; Cryptographic algorithms; Cryptographic attacks; Ciphers; Classical cryptography; Decipherment; Disk encryption; Cryptographic hardware; History of cryptography; Key management; Cryptography law; National Security Agency cryptography; Cryptography organizations; Cryptographic primitives; Cryptographic protocols; Public-key cryptography; Quantum cryptography; Security engineering; Cryptographic software; Cryptography standards; Steganography; Symmetric-key cryptography; Theory of cryptography. \\ \addlinespace[8pt] \midrule

\rowcolor{odysseyrow} \multicolumn{1}{c}{\textbf{\emph{(14) Puzzles}}} \\ \addlinespace[4pt]
Adventure game puzzles; Anagrams; Puzzle books; Chess problems; Puzzle competitions; Cryptography contests; Puzzle designers; Escape rooms; Guessing games; Locked-room mysteries; Puzzle magazines; Puzzle manufacturers; Mathematical puzzles; Mazes; Mechanical puzzles; Paradoxes; Riddles; Tiling puzzles; Unsolvable puzzles; Puzzle video games; Word puzzles; Logic puzzles; Nonograms; Rubik's Cube; Shunting puzzles; Sudoku; Combination puzzles; Jigsaw puzzles; Mechanical puzzle cubes; Chess composers; Fairy chess; Mathematical chess problems; Folklore featuring impossible tasks; Koan; Puzzle hunts; Alternate reality games; The Amazing Race; Crosswords; Word puzzle video games; Puzzle video games by series; Logical paradoxes; Mathematical paradoxes; Optical illusions; Hidden object games; Tile-matching video games; Sudoku video games; Maze games; Falling block puzzle games; Mahjong video games; Professor Layton; Puyo Puyo; Tetris; Tetris games; Puzzle Bobble; Bejeweled; Angry Birds; Portal (series); Myst games; Bomberman; Lemmings games; Brain teasers; Illusions; Auditory illusions; Tactile illusions; Dilemmas. \\ \addlinespace[8pt] \midrule

\rowcolor{odysseyrow} \multicolumn{1}{c}{\textbf{\emph{(15) Code}}} \\ \addlinespace[4pt]
Algorithms; Anti-patterns; Programming contests; Code refactoring; Concurrent computing; Programming constructs; Program derivation; Data structures; Debugging; Computer programming folklore; Programming games; Programming idioms; Programming language syntax; Programming language topics; Programming languages; Computer libraries; Live coding; Programming paradigms; Programming principles; Software optimization; Quantum programming; Self-hosting software; Software design; Software design patterns; Source code; Source code generation; Streaming algorithms; Programming tools; Visual programming; Computer programmers; Computer programming books; Computer programming stubs; Cryptography; Cryptography lists and comparisons; Cryptographic algorithms; Applications of cryptography; Cryptographic attacks; Ciphers; Classical cryptography; Cryptographers; Decipherment; Disk encryption; Cryptologic education; End-to-end encryption; Cryptographic hardware; History of cryptography; Key management; Cryptography law; National Security Agency cryptography; Cryptography organizations; Cryptographic primitives; Cryptographic protocols; Public-key cryptography; Cryptography publications; Quantum cryptography; Security engineering; Cryptographic software; Cryptography standards; Steganography; Symmetric-key cryptography; Theory of cryptography; Cryptography stubs; Computer programmers; Algorithms; Anti-patterns; Programming contests; Code refactoring; Concurrent computing; Programming constructs; Program derivation; Data structures; Debugging; DLL injection; Computer programming folklore; Programming games; Programming idioms; Programming interfaces; Programming language syntax; Programming language topics; Programming languages; Computer libraries; Live coding; Method (computer programming); Programming paradigms; Programming principles; Software optimization; Quantum programming; Self-hosting software; Software design; Software design patterns; Source code; Source code generation; Streaming algorithms; Programming tools; Visual programming.
\end{longtable}

\end{document}